\documentclass[11pt]{article}

\usepackage{etoolbox}
\newtoggle{arxiv}
\toggletrue{arxiv}

\usepackage{booktabs,amssymb,threeparttable}

\iftoggle{arxiv}{
\usepackage[sort&compress,numbers]{natbib}
}{}

\usepackage{boxedminipage}
\usepackage{multirow,nicefrac}
\usepackage{makecell,upgreek}
\usepackage{footnote}
\usepackage{longtable}
\usepackage{tablefootnote}
\usepackage[T1]{fontenc}
\usepackage{verbatim} 
\usepackage[utf8]{inputenc}

\usepackage{booktabs}
\usepackage{tabularx}
\usepackage{adjustbox}

\iftoggle{arxiv}{
\usepackage[margin=1in]{geometry}
}{}
\usepackage{algorithm}
\usepackage{algpseudocode}

\usepackage[dvipsnames]{xcolor}

\usepackage{graphicx}
\usepackage{caption}
\usepackage{subcaption}

\usepackage{colortbl}
\definecolor{lightgray}{gray}{0.9}  %

\usepackage{amsmath}
\usepackage{amsthm}
\iftoggle{arxiv}{
\usepackage[breaklinks=true]{hyperref}
}{}
\hypersetup{
	colorlinks=true,
	linkcolor=blue,
	citecolor=blue,
	urlcolor=blue
}

\usepackage[capitalise,nameinlink]{cleveref}

\usepackage{amsfonts,amssymb,mathrsfs}
\usepackage{mathtools}

\usepackage{appendix}

\theoremstyle{plain}
	\newtheorem{theorem}{Theorem}

	\newtheorem{proposition}{Proposition}

	\theoremstyle{definition}

	\newtheorem{definition}{Definition}

\Crefname{appendix}{Appendix}{Appendices}

\usepackage{autonum}

\usepackage[scaled]{beramono}
\usepackage{listings}
\definecolor{varorange}{RGB}{230,126,34}     %
\definecolor{opblue}{RGB}{52,152,219}        %
\definecolor{ifred}{RGB}{192,57,43}          %
\definecolor{commentgray}{RGB}{150,150,150}  %
\definecolor{codebg}{rgb}{0.98,0.98,0.98}    %

\newsavebox\codebox

\usepackage{authblk}

\newcommand{\ee}{\mathbb{E}}
\newcommand{\pp}{\mathbb{P}}

\newcommand{\Hnull}{\ensuremath{\mathbf{H}_{\mathbf{0}}}}
\newcommand{\Halt}{\ensuremath{\mathbf{H}_{\mathbf{A}}}}

\renewcommand{\epsilon}{\varepsilon}

\def\ddefloop#1{\ifx\ddefloop#1\else\ddef{#1}\expandafter\ddefloop\fi}
\def\ddef#1{\expandafter\def\csname bb#1\endcsname{\ensuremath{\mathbb{#1}}}}
\ddefloop ABCDEFGHIJKLMNOPQRSTUVWXYZ\ddefloop

\def\ddefloop#1{\ifx\ddefloop#1\else\ddef{#1}\expandafter\ddefloop\fi}
\def\ddef#1{\expandafter\def\csname frak#1\endcsname{\ensuremath{\mathfrak{#1}}}}
\ddefloop ABCDEFGHIJKLMNOPQRSTUVWXYZ\ddefloop

\def\ddefloop#1{\ifx\ddefloop#1\else\ddef{#1}\expandafter\ddefloop\fi}
\def\ddef#1{\expandafter\def\csname fr#1\endcsname{\ensuremath{\mathfrak{#1}}}}
\ddefloop ABCDEFGHIJKLMNOPQRSTUVWXYZ\ddefloop

\def\ddefloop#1{\ifx\ddefloop#1\else\ddef{#1}\expandafter\ddefloop\fi}
\def\ddef#1{\expandafter\def\csname eul#1\endcsname{\ensuremath{\EuScript{#1}}}}
\ddefloop ABCDEFGHIJKLMNOPQRSTUVWXYZ\ddefloop

\def\ddefloop#1{\ifx\ddefloop#1\else\ddef{#1}\expandafter\ddefloop\fi}
\def\ddef#1{\expandafter\def\csname scr#1\endcsname{\ensuremath{\mathscr{#1}}}}
\ddefloop ABCDEFGHIJKLMNOPQRSTUVWXYZ\ddefloop

\def\ddefloop#1{\ifx\ddefloop#1\else\ddef{#1}\expandafter\ddefloop\fi}
\def\ddef#1{\expandafter\def\csname b#1\endcsname{\ensuremath{\mathbf{#1}}}}
\ddefloop ABCDEFGHIJKLMNOPQRSTUVWXYZ\ddefloop

\def\ddefloop#1{\ifx\ddefloop#1\else\ddef{#1}\expandafter\ddefloop\fi}
\def\ddef#1{\expandafter\def\csname bhat#1\endcsname{\ensuremath{\hat{\mathbf{#1}}}}}
\ddefloop ABCDEFGHIJKLMNOPQRSTUVWXYZ\ddefloop

\def\ddefloop#1{\ifx\ddefloop#1\else\ddef{#1}\expandafter\ddefloop\fi}
\def\ddef#1{\expandafter\def\csname btil#1\endcsname{\ensuremath{\tilde{\mathbf{#1}}}}}
\ddefloop ABCDEFGHIJKLMNOPQRSTUVWXYZ\ddefloop

\def\ddefloop#1{\ifx\ddefloop#1\else\ddef{#1}\expandafter\ddefloop\fi}
\def\ddef#1{\expandafter\def\csname bst#1\endcsname{\ensuremath{\mathbf{#1}^\star}}}
\ddefloop ABCDEFGHIJKLMNOPQRSTUVWXYZ\ddefloop

\def\ddefloop#1{\ifx\ddefloop#1\else\ddef{#1}\expandafter\ddefloop\fi}
\def\ddef#1{\expandafter\def\csname bst#1\endcsname{\ensuremath{\mathbf{#1}^\star}}}
\ddefloop abcdeghijklmnopqrstuvwxyz\ddefloop

\def\ddefloop#1{\ifx\ddefloop#1\else\ddef{#1}\expandafter\ddefloop\fi}
\def\ddef#1{\expandafter\def\csname bhat#1\endcsname{\ensuremath{\hat{\mathbf{#1}}}}}
\ddefloop abcdefghijklmnopqrstuvwxyz\ddefloop

\def\ddefloop#1{\ifx\ddefloop#1\else\ddef{#1}\expandafter\ddefloop\fi}
\def\ddef#1{\expandafter\def\csname b#1\endcsname{\ensuremath{\mathbf{#1}}}}
\ddefloop abcdeghijklnopqrstuvwxyz\ddefloop

\def\ddefloop#1{\ifx\ddefloop#1\else\ddef{#1}\expandafter\ddefloop\fi}
\def\ddef#1{\expandafter\def\csname barb#1\endcsname{\ensuremath{\bar{\mathbf{#1}}}}}
\ddefloop abcdefghijklmnopqrstuvwxyz\ddefloop

\def\ddef#1{\expandafter\def\csname c#1\endcsname{\ensuremath{\mathcal{#1}}}}
\ddefloop ABCDEFGHIJKLMNOPQRSTUVWXYZ\ddefloop
\def\ddef#1{\expandafter\def\csname h#1\endcsname{\ensuremath{\widehat{#1}}}}
\ddefloop ABCDEFGHIJKLMNOPQRSTUVWXYZ\ddefloop
\def\ddef#1{\expandafter\def\csname hc#1\endcsname{\ensuremath{\widehat{\mathcal{#1}}}}}
\ddefloop ABCDEFGHIJKLMNOPQRSTUVWXYZ\ddefloop
\def\ddef#1{\expandafter\def\csname t#1\endcsname{\ensuremath{\widetilde{#1}}}}
\ddefloop ABCDEFGHIJKLMNOPQRSTUVWXYZ\ddefloop
\def\ddef#1{\expandafter\def\csname tc#1\endcsname{\ensuremath{\widetilde{\mathcal{#1}}}}}
\ddefloop ABCDEFGHIJKLMNOPQRSTUVWXYZ\ddefloop

\title{Are AI Watermarks Valid?}

\author[1]{Nghi Dao}
\author[2]{Xiaoyu Wu}
\author[2]{Zhiwei Steven Wu}
\author[1,3]{Adam Block}
\affil[1]{Department of Computer Science, Columbia University}
\affil[2]{Software and Societal Systems Department, Carnegie Mellon University}
\affil[3]{Department of Electrical Engineering, Columbia University}
\affil[ ]{\texttt{\{nmd2167, adam.block\}@columbia.edu}, \texttt{\{xiaoyuwu, zstevenwu\}@cmu.edu}}
\date{}

\usepackage{microtype}
\usepackage{enumitem}
\usepackage{amsmath,amssymb,amsthm,mathtools}
\usepackage{booktabs,multirow}

\usepackage{graphicx}
\usepackage{xcolor}
\usepackage{algorithm}
\usepackage{algpseudocode}

\newcommand{\FPR}{\mathrm{FPR}}
\newcommand{\TPR}{\mathrm{TPR}}

\begin{document}

\maketitle

\begin{abstract}
     Watermarking has emerged as a promising approach to identifying AI-generated content with statistically rigorous control over false-positive rates (FPR). In this work, we show that existing watermarks often fail to achieve their advertised control. First, many watermarking schemes \emph{guarantee validity only on average} over the random watermark `key', while deployment requires controlling false positives for the fixed key that is actually used. Second, while a common approach used in the literature is to employ conformal-based approaches to resolve this problem by using a held-out set to estimate a detection threshold and provide validity for a fixed key, this approach critically relies on \emph{exchangeability} between calibration and deployment data, which can fail even under seemingly benign distribution shifts. We show empirically that both issues lead popular schemes such as textGrain (currently deployed by OpenAI) and SynthID (currently deployed by Google and adapted by Anthropic) to fail to provide the intended FPR control, resulting in a significant rate of content incorrectly being flagged as AI-generated. To address the gap between claimed and actual FPR, we adopt the PAC-validity framework, which adjusts detection thresholds to control FPR with high probability over keys, at a small cost in TPR for most watermarks.  Taken together, our results suggest that the use of watermarks in high-stakes contexts where the cost of a false positive is high should be approached with caution and, in such settings, the community should move away from the conformalized statistical tests that are frequently deployed today in favor of the PAC-validity framework.\footnote{All code that can be used to reproduce our results is available at \url{https://github.com/Nghi-Dao/watermark_validity}.}

\end{abstract}

\section{Introduction}\label{sec:intro}

With the increasing quality and ubiquity of generative models, the ability to distinguish human content from that generated by AI has become a practical and legal necessity, with the EU AI Act recently mandating that all covered generative model providers employ machine-readable marking for detectability, i.e., watermarking \citep{euaiact2024,eucop2026}.  From early civil litigation cases surrounding accusations of AI-assisted cheating in schools~\citep{harris2024adams,gerezgiher2025yang,brown2026yangappeal,bleizeffer2025yale} to the identification of deepfakes in the media~\citep{chesney2019deepfakes,rose2026trust,nilssonjulien2026netanyahu}, the need for reliable detection of AI-generated content is clear.  Thus far, detection of AI-generated content has been approached in two ways: post-hoc detection and statistically valid \emph{watermarking}.  Post-hoc detection, e.g. Pangram and GPTZero~\citep{emi2024technicalreportpangramaigenerated,glickenhaus2026pangram4technicalreport,adam2026gptzerorobustdetectionllmgenerated}, trains a classifier on labeled data to achieve low error rates on evaluation datasets.  While extremely successful in many domains, post-hoc detectors do not come with false-positive rate (FPR) guarantees. For this reason, such detectors are insufficient for consequential applications where the cost of a false positive is high, such as those leading to accusations of academic misconduct, misidentification of authentic media, or undermining the interpretation of evidence.

\begin{figure}[t]
    \centering
    \includegraphics[width=1.0\linewidth]{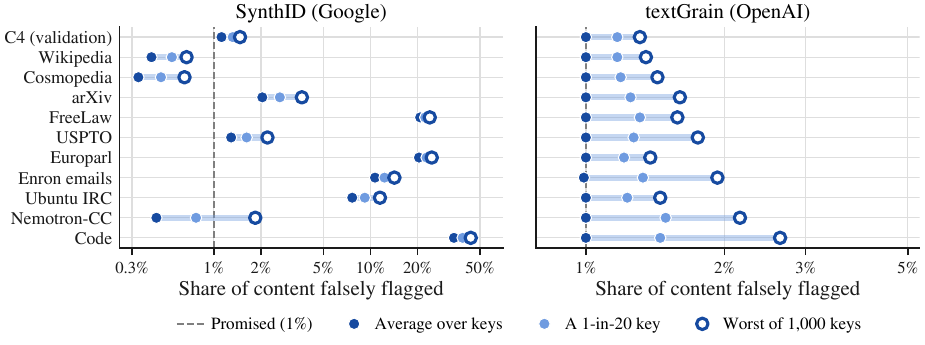}
\caption{\textbf{An advertised FPR of $1\%$ does not ensure control across datasets or individual keys.} Actual FPR of the two deployed text watermarks on each dataset, at an advertised rate of $1\%$. (Left)~SynthID's mean varies with the dataset. (Right)~textGrain's mean stays near $1\%$, but some keys significantly exceed it.}
    \label{fig:promise}
\end{figure}

\begin{figure}[t]
    \centering
    \includegraphics[width=1.0\linewidth]{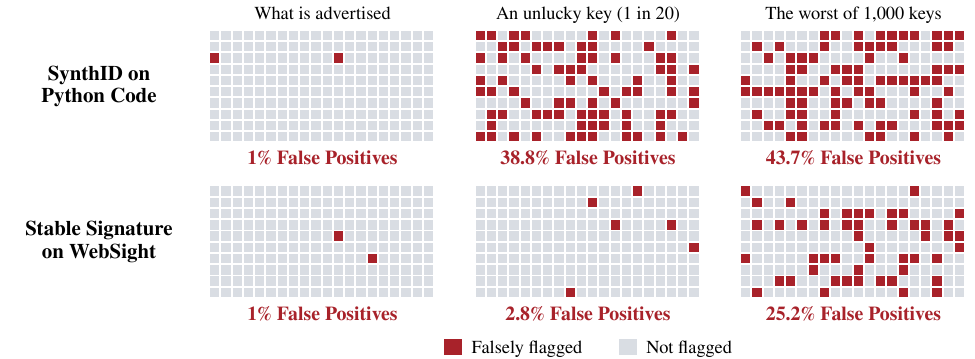}
\caption{\textbf{Actual FPR can exceed $40\%$ despite an advertised rate of $1\%$.} Each grid shows $200$ pieces of unwatermarked content, with the falsely flagged ones in red, at three rates: the advertised $1\%$, the 95th percentile across keys, and the maximum over $1{,}000$ keys. \textbf{Top (text):} SynthID on Python code. \textbf{Bottom (vision):} Stable Signature on screenshots of web pages (WebSight).}
    \label{fig:false_accusations}
\end{figure}

To provide a fully reliable guarantee in critical scenarios, watermarking, i.e., inserting a signal derived from a secret key into generated content, has emerged as a promising approach~\citep{aaronson2022watermarking,kirchenbauer2024watermarklargelanguagemodels,kuditipudi2024robustdistortionfreewatermarkslanguage,pmlr-v247-christ24a}.  A provider embeds a statistical signal during generation that follows a known distribution, and a detector can then test for the presence of this signal using formal statistical hypothesis testing to provide,  in theory, a statistical guarantee on the \emph{False Positive Rate} (FPR)~\citep{fernandez2023bricksconsolidatewatermarkslarge,huang2024optimalstatisticalwatermarking,Li_2025}, where the FPR is defined as the fraction of content incorrectly flagged as AI-generated. Since their introduction, watermarking schemes have moved from research to deployment, with Google and Anthropic using the SynthID-Text watermark in Gemini and Claude~\cite{anthropic2026watermark,dathathri2024scalable}, respectively, and OpenAI introducing textGrain~\cite{openai2026textgrain,openai2026textgrainreport} as an opt-in for API users and rolling it out to Codex in the European Union.  Beyond text domains, many image watermarking schemes have also been proposed and deployed~\citep{wen2023treeringwatermarksfingerprintsdiffusion,fernandez2023stablesignaturerootingwatermarks,yang2024gaussianshadingprovableperformancelossless,fernandez2024videosealopenefficient,gowal2025synthidimageimagewatermarkinginternet}.  

The omnipresence of watermarking in deployed systems and its potential legal and reputational implications make practical FPR controls of paramount importance.  
In this work, we revisit prior claims of statistical validity in watermarking schemes and show that they are often violated.  In particular, we ask the following question:
\begin{quotation}
    \emph{To what extent does the claimed control over false positive rates in existing watermarking schemes actually hold in practice, and what are the implications for their deployment?}
\end{quotation}
We show empirically that the \textbf{actual FPR of virtually all existing watermarking schemes can be substantially higher than advertised, even for schemes that claim theoretical validity}.  Underlying this disconnect between theory and practice, we identify two key issues: 
\begin{enumerate}
    \item the distinction between \emph{marginal} and \emph{conditional} validity, and
    \item the reliance of many watermarking schemes on an \emph{exchangeability} assumption that is often violated in practice.
\end{enumerate}

\textbf{Marginal and Conditional Validity.} Recall that the premise of statistical hypothesis testing is that observations are either drawn from a null distribution or an alternative distribution.  The goal is to develop a test with a controlled FPR $\alpha$ under the null, meaning an authentic observation will be incorrectly flagged as watermarked with probability at most $\alpha$.  Many current watermarks rely on the randomness of the watermark key itself: an unwatermarked document is unlikely to match a randomly drawn key, yielding a guarantee that averages over possible keys, which we refer to as \emph{marginal validity}.  In practice, a provider draws one key $\xi$ and uses it on all content. The FPR must be bounded for the given key, which we call \emph{conditional validity}.  While conditional validity implies marginal validity, the converse is not true: suppose $90\%$ of keys never flag human-written text, while the remaining keys falsely flag $10\%$ of it.  Then the conditional FPR of the $10\%$ of keys is $10\%$, yet the marginal FPR remains $1\%$.  A guarantee that holds over random keys can have substantially higher FPRs during deployment  (\Cref{fig:promise,fig:false_accusations}).
While many existing watermarking schemes theoretically demonstrate marginal validity, we show empirically that conditional validity often fails in practice. 

\textbf{Distribution Shift Eliminates Exchangeability.} To approximate the null distribution of a watermarking scheme whose exact test distribution may be hard to compute, several recent works~\cite{xie2025watermarkclassroomconformalframework,kuditipudi2024robustdistortionfreewatermarkslanguage}, including the version of SynthID-Text likely deployed in practice\footnote{While the precise watermarking details are not public, the results of \citet{dathathri2024scalable} primarily focus on their `Bayesian' version, which uses a learned `Bayesian’ score whose decision threshold is conformalized on unwatermarked text; this is precisely what we refer to here as the conformalized approach.}, use a Monte Carlo or conformalized approach. Such a scheme relies on a \emph{calibration set} of unwatermarked content to estimate the quantile of the watermarking test applied to a candidate text or image to determine how unlikely such a value would be under the null assumption that it is unwatermarked, and thus whether it should be flagged as AI-generated. This approach has become common throughout statistics due to its generality and minimal assumptions~\cite{angelopoulos2022gentleintroductionconformalprediction,Bates_2023,shafer2007tutorialconformalprediction}.  In fact, Monte Carlo tests require only \emph{exchangeability}, i.e., the calibration and test sets have to be sufficiently similar so that the calibration set's distribution is the same as that of content observed at deployment. However, we show empirically that this exchangeability assumption fails in practice. Conformalization holds on similar data distributions, but breaks under seemingly benign shifts, including future versions of a dataset and different content, such as Python Code, on which every key of the deployed SynthID detector fails (\cref{sec:conformal}; \cref{fig:bars_conformal,fig:time_series}).

\textbf{PAC-Validity.}
Having shown that neither marginal validity nor conformalized conditional validity sufficiently guarantees the low FPR required to deploy AI watermarking schemes, following \citet{gloaguen2026automarkenablingautoresearchdiscover}, we adopt a relaxation of conditional validity that can be certified.  Inspired by the Probably Approximately Correct (PAC) framework in learning theory~\cite{Valiant1984ATO}, we call a watermarking scheme \emph{$(\alpha', \delta)$-PAC valid} if, with probability at least $1 - \delta$ over the choice of the watermarking key, the detector is conditionally valid at level $\alpha'$; in words, this means that with high probability over the choice of key, we have chosen a `good' key that ensures the FPR is controlled.  For a given target level $\alpha$, we propose choosing $\alpha'$ and $\delta$ such that $\alpha' + \delta \leq \alpha$ so that a union bound ensures the overall false positive rate is well-controlled.  In this framework, we show the following result.

\begin{theorem}[Informal statement of \Cref{thm:markov}]\label{thm:pac_valid_informal}
    Any marginally valid watermarking scheme can be converted into a PAC-valid scheme by reducing the $p$-value threshold to $\alpha' \cdot \delta$.
\end{theorem} 
In particular, \Cref{thm:pac_valid_informal} suggests a natural, conservative approach to ensuring the FPRs of many existing watermarking schemes are well-controlled in practice: reduce the detection threshold, effectively requiring more evidence in order to flag a piece of content as AI-generated. We show empirically that this approach restores validity for all watermarking schemes that rely on marginal validity (\Cref{tab:summary}). We emphasize that \Cref{thm:pac_valid_informal} requires $p$-values that are marginally valid without assumptions on the content, so it does not apply to conformalized detectors such as the deployed SynthID-Text. 

\textbf{Recommendations.}
Our results expose severe failures in the statistical validity of existing watermarking schemes as deployed in practice and suggest that great care should be taken in the interpretation of positive detection, especially in high-stakes scenarios.  In particular, we recommend moving away entirely from conformalized watermarking schemes in domains where the costs of false positives are high.  While we show that PAC-validity applied to marginally valid schemes can be achieved in a general and rigorous way, it comes at the cost of reduced true positive rates and we hope that our work motivates future research into watermarking schemes that can achieve both high true positive rates and rigorous statistical validity in practice.

The remainder of the paper is organized as follows. We introduce the formal setup and definitions in \Cref{sec:prelims}, and show in \Cref{sec:marginal} that virtually all marginally valid image and text watermarks fail conditionally. We then show in \Cref{sec:conformal} that conformalized $p$-values fail in practice due to a lack of exchangeability; in particular, this suggests that conformalized detectors such as the one likely deployed in SynthID-Text \citep{dathathri2024scalable,gowal2025synthidimageimagewatermarkinginternet} can be invalid under distribution shift. In \Cref{sec:pac_validity}, we introduce the PAC-validity framework and show that correcting the detection threshold restores PAC-validity for every existing watermarking scheme with a marginally valid $p$-value. We discuss the implications and related work in \Cref{sec:related_work,sec:discussion}.

\section{Watermarking Setup: Definitions and Examples}\label{sec:prelims}

We begin by briefly reviewing statistical hypothesis testing, then frame watermarking as a hypothesis testing problem.  We then discuss conformalized, or Monte Carlo, estimated $p$-values and how they are used in contemporary watermarking applications, before detailing the watermarks studied.

\subsection{Watermarking and Hypothesis Testing}
\label{sec:testing}

We first recall the basic setup of statistical hypothesis testing, with details in standard references such as \citet{lehmann2022testing}.  We treat watermarking as a \emph{statistical hypothesis test}, following \citet{huang2024optimalstatisticalwatermarking,block2025gaussmark,zhao2026marktune}. Formally, in hypothesis testing, one strives to use observed data to distinguish between a null hypothesis $\Hnull$ and an alternative hypothesis $\Halt$, both of which are nonempty collections of probability distributions from one of which the observed data is sampled.  In the context of watermarking, the observed data consists of the pair $(\xi, X)$ where $\xi \in \Xi$ is a watermarking \emph{key} and $X \in \cX$ is a piece of content, e.g. some text or an image.  The null hypothesis consists of the case where $\xi$ and $X$ are \emph{independent}, meaning that the content was generated without the use of the watermarking key $\xi$, while the alternative hypothesis consists of the case where $\xi$ and $X$ are \emph{dependent}.  Typically, we assume knowledge of the distribution $\Pi$ over keys $\xi$ as well as the conditional distribution $Q_\xi$ of content $X$ given a key $\xi$.

\textbf{Watermarking.} A \emph{watermarking scheme} refers to the construction of the key distribution $\Pi$, the generative distribution $Q_\xi$, as well as a \emph{test} $\phi: \cX \times \Xi \to \left\{ 0,1 \right\}$, where $\phi(X, \xi) = 1$ indicates that the watermark is detected in content $X$ with key $\xi$. We say that a given test $\phi$ is \emph{valid} at level $\alpha$ if the FPR is controlled at level $\alpha$, i.e., if $\pp_{\Hnull}(\phi(X, \xi) = 1) \leq \alpha$; we say that a test has \emph{power} $\beta$ if the true-positive rate (TPR) is at least $\beta$, i.e., if $\pp_{\Halt}(\phi(X, \xi) = 1) \geq \beta$.  Moreover, a $p$-value associated with a set of tests $\phi_\alpha$ is a statistic $p: \cX \times \Xi \to [0,1]$ such that rejecting the null hypothesis when $p(X, \xi) \leq \alpha$ is equivalent to rejecting the null hypothesis with test $\phi_\alpha$; smaller $p$-values indicate stronger evidence against the null hypothesis. In the context of watermarking, we will denote by $p_\xi(x)$ the associated $p$-value for a given piece of content $x$ and key $\xi$ when the scheme itself is clear from context.

\textbf{Marginal and Conditional Validity.}  In the description above, we have defined validity in abstract statistical terms. But in practice, there are two distinct notions of validity, depending on whether the FPR is controlled on average over the choice of key $\xi$ or for a fixed key $\xi$.  More precisely, for a given key $\xi$ and distribution $P$ of unwatermarked content, we define the FPR and TPR of a watermarking scheme at level $\alpha$ as
\begin{align}\label{eq:fpr}
    \FPR_P(\xi; \alpha) = \pp_{X \sim P}(p_\xi(X) \leq \alpha) \quad \text{and} \quad \TPR(\xi; \alpha) = \pp_{X \sim Q_\xi}(p_\xi(X) \leq \alpha).
\end{align}
We can then formally define the three notions of validity as follows:
\begin{definition}[Marginal, Pointwise, and Conditional Validity]\label{def:validity}
    A watermarking scheme is \emph{marginally valid} at level $\alpha$ if $\ee_{\xi \sim \Pi}[\FPR_P(\xi; \alpha)] \leq \alpha$; it is \emph{pointwise valid} at level $\alpha$ if $\pp_{\xi \sim \Pi}(p_\xi(x) \leq \alpha) \leq \alpha$ for every input $x \in \mathcal{X}$; and it is \emph{conditionally valid} at level $\alpha$ if $\FPR_P(\xi; \alpha) \leq \alpha$ for $\xi \in \Xi$ with probability $1$.
\end{definition}
Both pointwise validity and conditional validity imply marginal validity, but the converse is not true in general.  Intuitively, to maintain FPR control under marginal validity, one is required to resample the key $\xi$ for each piece of content. But if conditional validity is satisfied, one can use a fixed key $\xi$ for all content. For obvious reasons, the latter is the more desirable property in practice.

\subsection{Calibrated \texorpdfstring{$p$}{p}-values}
\label{sec:calibration}

Typically, watermarking schemes use a family of tests $\phi_\alpha$, each valid at level $\alpha$ under a theoretical null distribution.  These tests are generally constructed using a test statistic whose distribution is known under the null hypothesis; this allows an exact $p$-value to be computed for each piece of content by evaluating the appropriate tail probability under the null distribution.  However, in practice, the theoretical null may not hold exactly, or the distribution of the test statistic under the null may be difficult to compute.  In these cases, calibration replaces the deployed key’s theoretical null with an empirical one. Given the key $\xi$ and a calibration sample $X_1, \dots, X_n$ of unwatermarked content, independent of $\xi$, let $p_i = p_\xi(X_i)$ and define the \emph{calibrated} (conformal) $p$-value
\begin{equation}
\label{eq:conformal}
  \hat p_\xi(x) \;=\; \frac{1 + \#\{i \le n : p_i \le p_\xi(x)\}}{n+1}.
\end{equation}
In words, the calibrated $p$-value is the fraction of calibration $p$-values that are less than or equal to the $p$-value of the content being tested, measuring how much more extreme a piece of content is than the calibration set.

In addition to the advantages of conformalized $p$-values in the case where the null distribution is difficult or impossible to compute, calibration has the added advantage that it  guarantees validity conditional on the deployed key instead of mere marginal validity. 
Unfortunately, to maintain any kind of validity, these calibrated $p$-values require \emph{exchangeability} between the unwatermarked content seen at deployment and the calibration sample~\citep{shafer2007tutorialconformalprediction,angelopoulos2022gentleintroductionconformalprediction}.  Intuitively, exchangeability requires that the calibration set and the deployment content be drawn from the same distribution; in this case, conditional validity is guaranteed~\citep{Bates_2023}.

\subsection{Watermarks Studied}
\label{sec:watermarks}

We investigate the statistical validity of seven text watermarking schemes and three image
watermarking schemes. We now summarize how each scheme
embeds and detects its watermark, together with its validity
guarantees and the assumptions underlying them; see also Table~\ref{tab:watermark-validity}.

\begin{table}[t]
    \centering
    \begin{threeparttable}
        \caption{\textbf{Validity guarantees of the evaluated watermark detectors.}
        }
        \label{tab:watermark-validity}
        \scriptsize
        \setlength{\tabcolsep}{2pt}
        \renewcommand{\arraystretch}{1.15}
        \newcommand{\wmhead}[1]{\begin{tabular}[t]{@{}c@{}}#1\end{tabular}}
        \newcommand{\wmcheck}{\ensuremath{\checkmark}}
        \begin{tabular*}{\linewidth}{@{\extracolsep{\fill}}l*{10}{c}@{}}
            \toprule
            & \multicolumn{7}{c}{\textit{Text watermarks}}
            & \multicolumn{3}{c}{\textit{Image watermarks}} \\
            \cmidrule(lr){2-8} \cmidrule(l){9-11}
            \textbf{Validity}
            & \textbf{KGW}
            & \textbf{TextSeal}
            & \textbf{\wmhead{Gumbel-\\max}}
            & \textbf{\wmhead{SynthID-\\Base}}
            & \textbf{SynthID}
            & \textbf{textGrain}
            & \textbf{\wmhead{Gauss-\\Mark}}
            & \textbf{\wmhead{Tree-\\Ring}}
            & \textbf{\wmhead{Stable\\Signature}}
            & \textbf{VideoSeal} \\
            \midrule
            \addlinespace[2pt]
            Exact marginal
            & & & & & & & \wmcheck & & \wmcheck & \wmcheck \\
            \addlinespace[3pt]
            Approx.\ marginal
            & \wmcheck & \wmcheck & \wmcheck & \wmcheck
            & & \wmcheck & & & & \\
            \addlinespace[3pt]
            Conditional
            & & & & & \wmcheck\tnote{a} & & & \wmcheck\tnote{b} & & \\
            \addlinespace[2pt]
            \bottomrule
        \end{tabular*}
        \begin{tablenotes}[flushleft]
            \footnotesize
            \item[a] Assuming the input distribution is exchangeable with the calibration data.
            \item[b] Under the assumed Gaussian null for the recovered coefficients.
        \end{tablenotes}
    \end{threeparttable}
\end{table}

\medskip
\noindent\textbf{Text watermarks.}
\begin{enumerate}
    \item \textbf{KGW}~\citep{kirchenbauer2024watermarklargelanguagemodels}.
    KGW promotes a green list of tokens determined by the key
    and preceding token, and detects the watermark by counting
    green tokens among distinct (context, token) pairs. We use the
    exact binomial tail rather than the original $z$-test.
    \emph{Validity:} marginal validity, assuming the green lists of
    distinct contexts are independent.

    \item \textbf{Gumbel-max}~\citep{aaronson2022watermarking}.
    Gumbel-max sampling selects tokens by using the key to sample uniform random variables $u_t$ based on the previous token, then combines these with the model's output distribution to sample the next token. Detection compares the sum of $-\log(1-u_t)$ over the selected tokens with a Gamma null. \emph{Validity:} approximate marginal validity, relying on independence across scored contexts. 

    \item \textbf{TextSeal}~\citep{sander2026textseallocalizedllmwatermark}.
    TextSeal is built on Gumbel-max, using two keys for sampling and detection. Since Gumbel-max samples based on the previous tokens, it is more likely to produce common phrases. TextSeal randomly switches between the two keys, ensuring output diversity.

    \item \textbf{SynthID-Text}~\citep{dathathri2024scalable}.
    SynthID-Text uses tournament sampling to favor tokens with larger key-dependent $g$-values. We study two detectors. \emph{SynthID-Base}, the mean score with a normal $p$-value, has \emph{approximate marginal validity}. \emph{SynthID}, the Bayesian detector used in practice, has no analytic null, so we calibrate it for each key  (Section~\ref{sec:calibration}). It is \emph{conditionally valid} only on average over the calibration sample, and only if calibration and test content are exchangeable.

    \item \textbf{textGrain}~\citep{openai2026textgrain}.
    textGrain biases token sampling using key-dependent costs under an entropy budget. Detection sums scores derived from these costs and compares the sum with a Gamma null distribution. As OpenAI's detector is not public, we reimplement it from its technical report~\cite{openai2026textgrainreport}. \emph{Validity:} approximate marginal validity, relying on independence across scored contexts.

    \item \textbf{GaussMark}~\citep{block2025gaussmark}.
    GaussMark adds Gaussian noise $\sigma Z$ to a model weight matrix, with $Z$ serving as the key. Detection uses the normalized projection of the unperturbed model's log-likelihood gradient onto $Z$. \emph{Validity:} exact marginal validity, since the statistic has a Gaussian distribution over the key for every fixed input.
    \end{enumerate}

\medskip
\noindent\textbf{Image watermarks.}
\begin{enumerate}
    \item \textbf{Tree-Ring}~\citep{wen2023treeringwatermarksfingerprintsdiffusion}.
    Tree-Ring embeds a key pattern in the Fourier transform of a diffusion model's initial noise. Detection inverts the diffusion process and compares the recovered coefficients with the key. \emph{Validity:} conditional validity, but only if the recovered coefficients follow the assumed Gaussian null.

    \item \textbf{Stable Signature}~\citep{fernandez2023stablesignaturerootingwatermarks}.
    Stable Signature fine-tunes a diffusion model's latent decoder to embed a fixed $k$-bit signature. Detection extracts the bits and tests their agreement with the signature using a binomial tail. \emph{Validity:} exact marginal validity for a uniformly random signature independent of the extractor.

    \item \textbf{VideoSeal}~\citep{fernandez2024videosealopenefficient}.
    VideoSeal uses a trained embedder and extractor to insert and recover a $k$-bit message, which serves as the key. We evaluate its image functionality using a binomial test of the number of matching bits. \emph{Validity:} exact marginal validity for a uniformly random key independent of the extractor.
\end{enumerate}

\section{Validity Over Keys Does Not Imply Validity Given a Key}
\label{sec:marginal}

\begin{figure}[t]
    \centering
    \includegraphics[width=\linewidth]{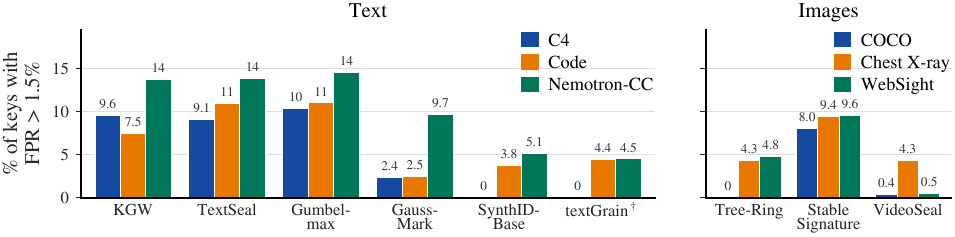}
\caption{\textbf{Under the theoretical cutoff, many keys greatly exceed the target FPR.} Percentage of $1{,}000$ keys whose FPR exceeds $1.5\%$ at the theoretical cutoff $p_\xi(x)\le\alpha$ with $\alpha=1\%$. (Left)~Text watermarks on C4 validation, Python code, and Nemotron-CC. (Right)~Image watermarks on held-out COCO, chest X-rays, and WebSight. $^\dagger$Our reimplementation.}
    \label{fig:bars_marginal}
\end{figure}

In this section, we investigate the extent to which the difference between marginal and conditional validity (\Cref{def:validity}) matters in practice.  Every watermarking scheme we study, with the exception of SynthID (for text) and Tree-Ring (for images), has a marginally valid $p$-value, at least approximately.  As we stated above, however, marginal validity is not a practically relevant notion, as providers deploy a single key and desire controlled false positive rates for that key on a large amount of content.  In \Cref{sec:exp-setup}, we describe our experimental setup, and in \Cref{sec:standard} we show that for most schemes, the deployment setting frequently leads to substantially higher FPRs than advertised.

\subsection{Experimental setup}
\label{sec:exp-setup}

\begin{figure}[t]
    \centering
    \includegraphics[width=\linewidth]{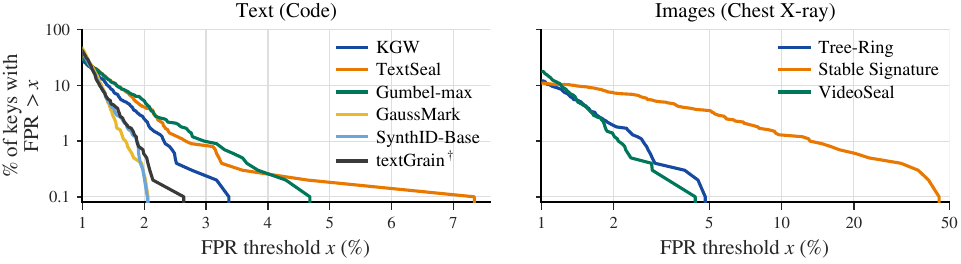}
\caption{\textbf{Some keys have FPRs several times the target.} Share of the $1{,}000$ keys whose FPR exceeds $x$ under the theoretical cutoff at $\alpha=1\%$, for (Left)~the text watermarks on Code and (Right)~the image watermarks on Chest X-ray. 
$^\dagger$Our reimplementation.}
    \label{fig:fpr_distribution}
\end{figure}

We consider 12 commonly used text corpora and 8 image datasets, which we summarize in \Cref{tab:datasets}.  For each dataset, we use $10{,}000$ sequences (or images) unless otherwise noted, and for each of $1{,}000$ randomly sampled keys, we compute the FPR at a target $\alpha = 1\%$ using the theoretical $p$-value for each watermarking scheme.  To simplify the presentation, we restrict the focus in the main body of the paper to three text datasets (C4 \citep{raffel2023exploringlimitstransferlearning}, Nemotron-CC \citep{diao2025climb}, and source-code \citep{codeparrot2022clean}) and three image datasets (COCO \citep{lin2015microsoftcococommonobjects}, chest X-rays \citep{kermany2018}, and WebSight \citep{laurençon2024unlocking}). Only GaussMark and Tree-Ring need the generative model during detection. The other detectors need only the key (and a tokenizer for text) except for StableSignature and VideoSeal which also require a learned extractor. In all cases, we use the watermarking hyperparameters recommended by the authors of each scheme.

For each key-dataset pair, we measure the FPR at the theoretical cutoff $p_\xi(x) \le \alpha$ and call a key \emph{bad} if this rate exceeds $1.5\alpha$.  Ablations for different multiplicative thresholds and choices of $\alpha$ can be found in \cref{tab:marginal_text,tab:marginal_image,tab:conformal_text,tab:conformal_image,fig:alpha_grid_text,fig:alpha_grid_image}.  We emphasize that the large volume of samples ($10{,}000$ per dataset) ensures that the probability that a key exceeds $1.5\alpha$ due to sampling noise is less than $0.1\%$.

\subsection{Conditional validity fails for standard watermarks}
\label{sec:standard}

We find that marginal validity in theory does not translate into conditional validity in practice.  In other words, for all watermarking schemes studied, a substantial fraction of watermarking keys have FPRs on the studied datasets that greatly exceed the advertised FPR.
\Cref{fig:bars_marginal}(left) shows that for KGW, TextSeal, and Gumbel-max, around $10\%$ of keys are `bad' (meaning have FPR $>1.5\%$) on C4 and Code, and even worse on Nemotron-CC. To reiterate, these observations are consistent with marginal validity; indeed, the FPR when averaged over the keys remains below the target. Hence, if conditional validity is not considered, all three of these methods will be certified despite a high failure rate. GaussMark illustrates this point most clearly: its statistic is exactly Gaussian over the key for every fixed input (\cref{sec:watermarks}); its mean FPR is $1.00\%$. It is pointwise valid, yet about $10\%$ of keys are `bad' on Nemotron-CC. This observation holds for all studied watermarks on many domains; e.g., in \Cref{fig:promise}(right), OpenAI's textGrain demonstrates that the deployed keys can have FPRs that are $>2\times$ the advertised FPR on multiple datasets.  We further illustrate this point in \Cref{fig:fpr_distribution}, which shows the share of keys whose FPR exceeds a given threshold, demonstrating that the choice of $1.5\alpha$ is not in any way distinguished.  To conclude, marginal validity does not imply validity for the deployed key.

\begin{figure}[t]
    \centering
    \includegraphics[width=0.85\linewidth]{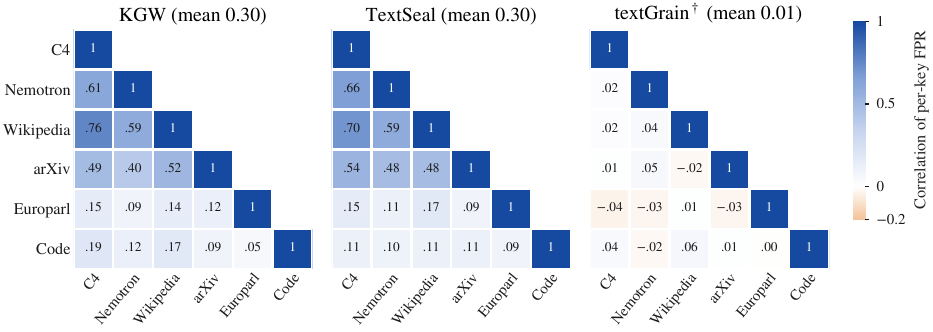}
\caption{\textbf{Per-key FPRs correlate across datasets for KGW and TextSeal but are nearly uncorrelated for textGrain.} Pairwise correlations, across $1{,}000$ keys, between the FPRs measured on different text datasets at $\alpha=1\%$ under the theoretical cutoff. $^\dagger$Our reimplementation.}
    \label{fig:fpr_correlation_main}
\end{figure}

\paragraph{What makes a key `bad?'}
Unwatermarked passages share structure, such as common phrases for most text watermarks and a common gradient direction for GaussMark. A fixed key interacts with this shared structure in the same way on every passage, which shifts its null distribution as a whole. For example, KGW decides which tokens are ``green'' using the previous tokens, and words like ``the'' commonly precede many words. A key whose green list contains common pairs of words will give more green tokens to the inputs, and so flags human-written text more often. The shift averages out over keys but not for a fixed key. This also implies that a key that performs poorly on one dataset is likely to perform poorly on another, where key performance across datasets is correlated. \Cref{fig:fpr_correlation_main} shows that KGW’s and TextSeal’s per-key FPRs are correlated across datasets, whereas textGrain’s are not. For textGrain, like SynthID, both of which use the four preceding tokens instead of a single one, the same context does not appear often enough to shift a key's FPR, except in repetitive text such as code. To test this, we varied the number of preceding tokens used by KGW and Gumbel-max. While using four tokens does remove bad keys for C4, it fails to remove them entirely for Code (\cref{tab:context-length,fig:context_sweep}). Longer contexts also make the watermark less robust, as it is easier to remove the watermark by editing a few tokens~\citep{kirchenbauer2024reliabilitywatermarkslargelanguage,jovanović2024watermarkstealinglargelanguage}. Further analysis of the mechanism by which keys become bad (including in the case of image watermarks) is provided in \cref{app:bad-keys}.

\section{Conformalization Does Not Survive Distribution Shift}
\label{sec:conformal}

The correlation of per-key FPRs across datasets (\cref{fig:fpr_correlation_main}) suggests that choosing a threshold to get a specific FPR using \cref{eq:conformal} on one dataset partially carries over to another. This is valid for a given key if the calibration and test data are exchangeable (\cref{sec:calibration}), and it is how the released SynthID detector works~\citep{dathathri2024scalable}. 

\begin{figure}[t]
    \centering
    \includegraphics[width=\linewidth]{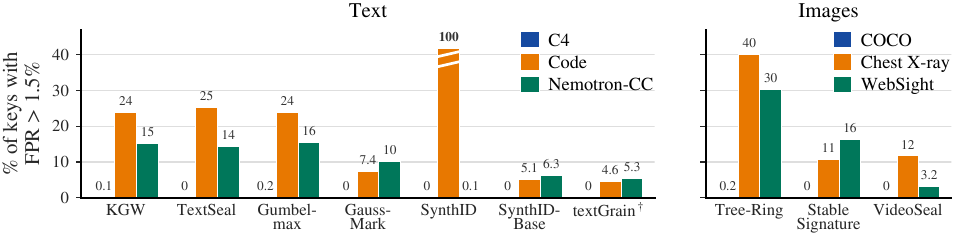}
\caption{\textbf{Conformalization can fail under distribution shift.} Percentage of $1{,}000$ keys whose FPR exceeds $1.5\%$ after per-key conformalization at target $\alpha=1\%$. (Left)~Text watermarks, conformalized on $10{,}000$ C4 training passages and evaluated on C4 validation, Python code, and Nemotron-CC. (Right)~Image watermarks, conformalized on $10{,}000$ COCO images and evaluated on held-out COCO, chest X-rays, and WebSight. $^\dagger$Our reimplementation.}
    \label{fig:bars_conformal}
\end{figure}

\subsection{Experimental setup}
\label{sec:conformal-setup}

We use the same keys, datasets, target level $\alpha = 1\%$, and definition of a bad key as in \cref{sec:exp-setup}, but replace the theoretical cutoff with a conformal cutoff from \Cref{eq:conformal}. We first use a held-out split of COCO and C4, each with $10{,}000$ samples, as calibration, then test each watermark on another split of the dataset (\Cref{tab:datasets}). Additionally, we study the change in FPR over time from the same source by conformalizing each key on one year of data and testing it on a later year, using FineWeb for text and Wikimedia Commons photos for images, both from 2013 to 2025.

\begin{figure}[t]
    \centering
    \includegraphics[width=\linewidth]{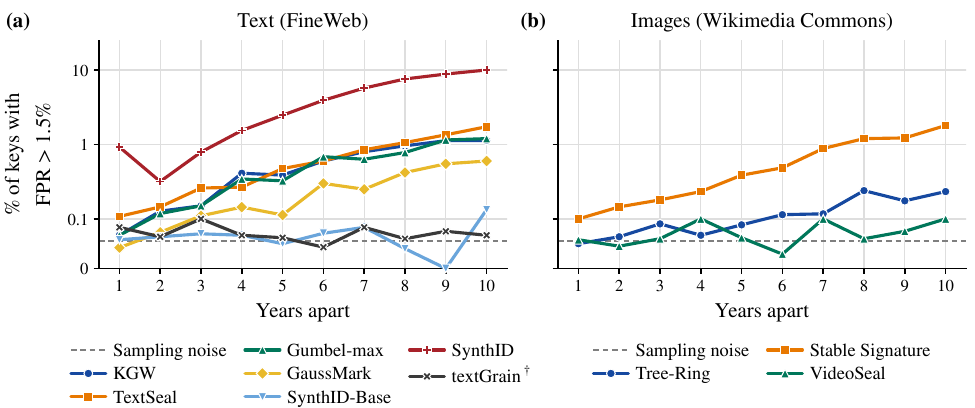}
\caption{\textbf{Conformalization can fail as deployment data change over time.} Share of the $1{,}000$ keys whose FPR after conformalization exceeds $1.5\%$ at target $\alpha=1\%$, when each key is conformalized on one year and tested on a later year of the same source: (a)~FineWeb for text and (b)~photos uploaded to Wikimedia Commons for images. Results are averaged over the year pairs with the same gap. $^\dagger$Our reimplementation.}
    \label{fig:time_series}
\end{figure}

\subsection{Exchangeability is required}

On C4 validation and held-out COCO, which are exchangeable with the calibration split, almost no key of any detector is bad after conformalization (\cref{fig:bars_conformal}), including the ones that were bad under the theoretical cutoff (\cref{fig:bars_marginal}).

\paragraph{Performance does not transfer to other datasets.}
On the other datasets, conformalization helps little and can increase the FPR on distributionally different datasets. On Nemotron-CC, whose FPRs are correlated with C4, the share of bad keys barely changes. On code, whose FPRs are only weakly correlated with C4, and on Chest X-rays, calibration makes things worse than the theoretical cutoff. For example, for Tree-Ring, conformalizing the keys leads to almost a ten-fold increase in bad keys (\cref{fig:bars_marginal,fig:bars_conformal}).

As discussed in \cref{sec:standard}, shared structure in images and human text introduces a bias during detection, which causes bad keys. Conformalizing a key on C4 corrects that key's bias on C4. On another dataset, the difference between the bias of the keys on C4 vs. those datasets remains. When the two shifts are weakly related, as for C4 and code (\cref{fig:fpr_correlation_main}), this difference is larger than the original bias, and conformalization creates more bad keys than it removes. Even for Nemotron-CC, whose per-key FPRs are clearly correlated with those on C4, the shifts are not similar enough.

\paragraph{Unaffected watermarks.}
SynthID-Base and textGrain have few bad keys on most datasets, with or without conformalization. Their FPRs do not correlate across datasets (\cref{fig:fpr_correlation_main,fig:fpr_correlation}); conformalization generally does not worsen performance but also does not improve it.

\paragraph{SynthID's detector fails.}
The Bayesian detector has no theoretical null, so conformalization is the only way to set its threshold. Conformalized on C4, it has no bad keys on C4 validation, but every key is bad on code (\cref{fig:bars_conformal}), with the average key incorrectly flagging over a third of unwatermarked text (\cref{tab:conformal_text}). Additionally, because there is no available analytic null distribution, the calibration amplifies the effect of the bias. We can see this in \cref{tab:conformal_text}, where SynthID has approximately 0\% bad keys on roughly half of the datasets and $100\%$ bad keys on the other half, with practically nothing in between.

\paragraph{Conformalization also degrades over time.}
Even when the calibration and test data come from the same source, the share of bad keys for some watermarks drifts over time. \Cref{fig:time_series} shows that for KGW, TextSeal, and Gumbel-max, and to a lesser extent GaussMark, the share of bad keys grows steadily with the number of years between calibration and testing. SynthID's Bayesian detector degrades the fastest, and Stable Signature shows the same trend for images. SynthID-Base, textGrain, Tree-Ring, and VideoSeal stay close to the noise level.

\subsection{Implications}
\label{sec:conformal-implications}

Our test sets are public corpora, and the differences between them are modest compared with what a deployed detector sees. A deployed detector is used for essays, news, code, reviews, and translations, written by different people, in different languages, and at different times, and the text it sees drifts over time~\citep{thomas2026llmevaluationforecastingproblem}. No calibration set is exchangeable with all of these. Our results show that the guarantee conformalization provides does not transfer even among public corpora. A guarantee for the deployed key must hold without exchangeability with the deployment data.

\section{PAC-Validity: Almost Every Key is Valid}\label{sec:pac_validity}

\begin{figure}[t]
    \centering
    \includegraphics[width=0.9\linewidth]{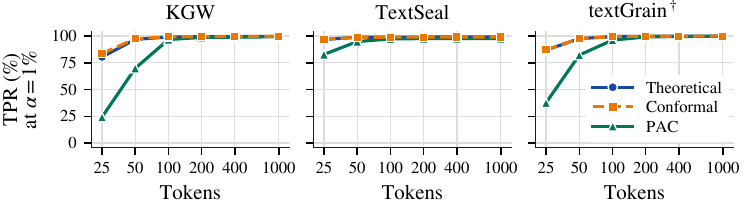}
\caption{\textbf{The PAC cutoff costs more detection power on short passages.} TPR against passage length for KGW, TextSeal, and textGrain$^\dagger$. At target $\alpha=1\%$, Theoretical uses $p_\xi(x)\le\alpha$, Conformal uses per-key calibration on C4 training data, and PAC uses $p_\xi(x)\le\alpha^2/4$.$^\dagger$Our reimplementation.}
    \label{fig:tpr_length}
\end{figure}

\begin{table}[t]
    \centering
    \begin{threeparttable}
        
\caption{\textbf{\boldmath FPR and detection power
at $\alpha = 1\%$.}
Each FPR entry is the largest value across all datasets in
\cref{tab:marginal_text,tab:marginal_image,tab:conformal_text,tab:conformal_image}.
\emph{Mean FPR} averages the FPR over $1{,}000$
keys under the theoretical cutoff; \emph{Keys with FPR $> 1.5\%$}
reports the percentage of these keys whose FPR exceeds $1.5\%$.
\emph{TPR} is the percentage of watermarked samples correctly detected.
The detection cutoffs are \emph{Marg.}, $p_\xi(x) \le \alpha$;
\emph{Conf.}, a per-key cutoff conformalized on C4 train
(text) or COCO (images) (\cref{sec:conformal}); and \emph{PAC},
$p_\xi(x) \le \alpha^2/4$ (\cref{thm:markov}).
Dashes indicate unavailable results; SynthID has no theoretical
$p$-value. $^\dagger$Our reimplementation.}
        \label{tab:summary}

        \footnotesize
        \setlength{\tabcolsep}{3pt}
        \renewcommand{\arraystretch}{1.1}
        \newcommand{\paccell}[1]{\cellcolor{black!5}#1}

        \begin{tabular*}{\linewidth}{@{\extracolsep{\fill}}l*{7}{c}}
            \toprule
            & \multicolumn{1}{c}{Mean FPR (\%)}
            & \multicolumn{3}{c}{Keys with FPR $> 1.5\%$ (\%)}
            & \multicolumn{3}{c}{TPR (\%)} \\
            \cmidrule(lr){2-2}
            \cmidrule(lr){3-5}
            \cmidrule(l){6-8}
            \textbf{Method}
            & \multicolumn{1}{c}{\textbf{Marg.}}
            & \multicolumn{1}{c}{\textbf{Marg.}}
            & \multicolumn{1}{c}{\textbf{Conf.}}
            & \multicolumn{1}{c}{\textbf{PAC}}
            & \multicolumn{1}{c}{\textbf{Marg.}}
            & \multicolumn{1}{c}{\textbf{Conf.}}
            & \multicolumn{1}{c}{\textbf{PAC}} \\
            \midrule

            \multicolumn{8}{@{}l}{\textit{Text watermarks}} \\
            KGW~\cite{kirchenbauer2024watermarklargelanguagemodels}
            & 0.90 & 14.7 & 23.8 & \paccell{0.0}
            & 99.8 & 99.8 & \paccell{99.3} \\

            TextSeal~\cite{sander2026textseallocalizedllmwatermark}
            & 0.90 & 14.2 & 25.6 & \paccell{0.0}
            & 99.1 & 99.1 & \paccell{97.6} \\

            Gumbel-max~\cite{aaronson2022watermarking}
            & 0.95 & 14.7 & 25.9 & \paccell{0.0}
            & 98.6 & 98.6 & \paccell{95.3} \\

            GaussMark~\cite{block2025gaussmark}
            & 1.00 & 17.1 & 16.7 & \paccell{0.0}
            & 99.4 & 99.3 & \paccell{89.4} \\

            SynthID-Base~\cite{dathathri2024scalable}
            & 0.99 & 5.1 & 6.3 & \paccell{0.0}
            & 97.8 & 97.8 & \paccell{93.4} \\

            SynthID~\cite{dathathri2024scalable}
            & -- & -- & 100.0 & \paccell{--}
            & -- & 100.0 & \paccell{--} \\

            textGrain$^\dagger$~\cite{openai2026textgrainreport}
            & 1.00 & 4.5 & 5.3 & \paccell{0.0}
            & 99.9 & 99.9 & \paccell{99.9} \\
            \midrule

            \multicolumn{8}{@{}l}{\textit{Image watermarks}} \\
            Tree-Ring~\cite{wen2023treeringwatermarksfingerprintsdiffusion}
            & 0.48 & 4.8 & 40.1 & \paccell{0.0}
            & 100.0 & 100.0 & \paccell{99.9} \\

            Stable Signature~\cite{fernandez2023stablesignaturerootingwatermarks}
            & 0.70 & 11.9 & 26.4 & \paccell{0.0}
            & 100.0 & 100.0 & \paccell{99.9} \\

            VideoSeal~\cite{fernandez2024videosealopenefficient}
            & 0.73 & 4.3 & 11.7 & \paccell{0.0}
            & 100.0 & 100.0 & \paccell{100.0} \\
            \bottomrule
        \end{tabular*}
    \end{threeparttable}
\end{table}

All current watermarks of which we are aware either rely on marginal validity or apply conformalized cutoffs.  Unfortunately, absent knowledge of the deployment distribution, neither of these approaches suffices to verify the \emph{conditional} validity that is desired.  In this section, following \citet{gloaguen2026automarkenablingautoresearchdiscover} and motivated by the classical theory of PAC learning \citep{Valiant1984ATO}, we thus propose a relaxation of conditional validity, \emph{PAC-validity}, that is achievable without knowledge of the deployment distribution.

\begin{definition}[PAC-validity]
\label{def:pac}
Let $\alpha', \delta \in [0,1]$. A detector with threshold $t$ is \emph{$(\alpha', \delta)$-PAC valid} under $P$ if
\[
  \pp_{\xi \sim \Pi}\bigl(\FPR_P(\xi; t) > \alpha'\bigr) \;\le\; \delta .
\]
It is \emph{valid at target $\alpha$} if it is $(\alpha', \delta)$-PAC valid for some $\alpha' + \delta \le \alpha$.
\end{definition}
In words, a key is `good' if its FPR is at most $\alpha'$, and a detector is PAC valid if the probability of drawing a bad key is at most $\delta$ for some small $\delta$.  Critically, if $\alpha' + \delta \leq \alpha$, then a union bound implies that the probability of drawing a bad key or drawing a good key and falsely flagging unwatermarked content is at most $\alpha$.  As $\delta \downarrow 0$, we recover conditional validity, which is the skyline of validity that we would like to achieve.  With this relaxation, we can now guarantee that most keys are valid without any knowledge of the deployment distribution as long as a watermark satisfies marginal validity.

\begin{theorem}
\label{thm:markov}
If a detector is marginally valid under data distribution $P$ at level $\alpha'\delta$, then the detector run at level $\alpha'\delta$
is $(\alpha', \delta)$-PAC valid under $P$. If it is pointwise valid, this holds under every $P$.
\end{theorem}

The proof is Markov's inequality applied to $\FPR_P(\xi; \alpha'\delta)$ over the key (\cref{app:markov}). A similar adjustment was suggested in \citet{gloaguen2026automarkenablingautoresearchdiscover} in the context of automated watermark discovery (cf. \Cref{sec:related_work}). The appeal of \Cref{thm:markov} is that it requires neither calibration data nor assumption on the content, so distribution shift cannot break it.  Moreover, without further assumptions the guarantee cannot be improved: for every level $t > \alpha'\delta$ there is an exactly pointwise valid scheme that is not $(\alpha', \delta)$-PAC valid at level $t$ (\cref{thm:tight}). With $\alpha' = \delta = \alpha/2$, the largest level allowed by $\alpha' + \delta \le \alpha$, the detector is run at $\alpha^2/4$, which is $0.0025\%$ for $\alpha = 1\%$. More generally, if we require PAC-validity to hold for multiple \emph{distinct} distributions conditionally on said distributions, we could further reduce the allowed level.

\Cref{thm:markov} requires a $p$-value that is valid without assumptions on the content, so it does not apply to SynthID, whose $p$-value is valid only under exchangeability. PAC-validity could also be impractical under conformalization as such a $p$-value cannot be smaller than $1/(n+1)$; thus, testing at $\alpha^2/4$ would require at least $40{,}000$ calibration passages per key.  

\paragraph{PAC-validity restores FPRs at the cost of power.}
\Cref{tab:summary} compares the three cutoffs. Under the PAC cutoff of \cref{thm:markov}, no key of any watermark is bad, including code and chest X-rays, where both the theoretical and the conformalized cutoffs fail. Unlike conformalization, this guarantee does not need a calibration set that resembles the deployment data, and it holds at every input length (\cref{fig:badkeys_length}). The price is detection power, since the PAC threshold is much stricter than $\alpha$. One interpretation of this threshold is that it is being more conservative on the promised FPR. For most watermarks, the TPR on images and $1{,}000$-token text input barely changes (refer to \cref{app:setup-tpr} for experimental details). The cost grows as passages get shorter, since a short passage carries less evidence of the watermark (\cref{fig:tpr_length,fig:tpr_length_appendix}). Part of the difference between text watermarks comes from their different generators: when all of them generate with the same model, the cost on $1{,}000$-token passages is small for every scheme (\cref{fig:tpr_llama8b}).

\section{Related Work}\label{sec:related_work}

We now discuss some related work on watermarking, as well as emphasizing the most closely related work, AutoMark~\citep{gloaguen2026automarkenablingautoresearchdiscover}.  We also discuss the broader literature on attacks and post-hoc detection of AI-generated content.

\paragraph{AI Content Detection.}
Classical digital watermarks hide a signal in an existing image, audio file, or video so that its origin can be proven later, using a secret key and a statistical test~\citep{650120}; \citet{aaronson2022watermarking} proposed moving the watermark into generation by introducing biases into the sampling process using a key and detecting them with a statistical test~\citep{aaronson2022watermarking,kirchenbauer2024watermarklargelanguagemodels}. Later schemes made the watermark distortion-free~\citep{kuditipudi2024robustdistortionfreewatermarkslanguage}, and undetectable without access to the key~\citep{pmlr-v247-christ24a,sander2026textseallocalizedllmwatermark}, and eventually deployed them~\citep{dathathri2024scalable,anthropic2026watermark,openai2026textgrainreport}. GaussMark~\citep{block2025gaussmark} instead embeds the key in the weights of the model. MarkTune~\citep{zhao2026marktune} builds on this by fine-tuning the model to strengthen this signal. Image watermarks fine-tune the latent decoder of a diffusion model~\citep{fernandez2023stablesignaturerootingwatermarks}, write the key into the noise used for generation~\citep{wen2023treeringwatermarksfingerprintsdiffusion,yang2024gaussianshadingprovableperformancelossless}, or embed a message with a learned encoder after generation~\citep{fernandez2024videosealopenefficient,gowal2025synthidimageimagewatermarkinginternet}. Of these, GaussMark, MarkTune, and Stable Signature are open-weight watermarks, since the watermark is embedded directly into the model weights and cannot be easily disabled. These works evaluate false positives by averaging over keys or with a single key on a single dataset.

\paragraph{Post-hoc detection.}
Other approaches detect AI-generated text without a watermark, either with a classifier trained on human and AI-written text, as in GPTZero~\citep{adam2026gptzerorobustdetectionllmgenerated} and Pangram~\citep{emi2024technicalreportpangramaigenerated,glickenhaus2026pangram4technicalreport}, or with a zero-shot classification using a language model's output~\citep{mitchell2023detectgptzeroshotmachinegeneratedtext,hans2024spottingllmsbinocularszeroshot}. These detectors do not require the model provider to embed a watermark, but also provide no FPR guarantees. Like conformalized watermark detectors (\cref{sec:conformal}), their scores are calibrated on held-out data~\citep{glickenhaus2026pangram4technicalreport,adam2026gptzerorobustdetectionllmgenerated}, and their FPRs are measured on a held-out test set, which says little about content unlike that data. Their own reports show higher FPRs on such content: Pangram 4 flags some human-written peer reviews polished by AI as AI-generated~\citep{glickenhaus2026pangram4technicalreport}, and GPTZero's FPR is highest on short documents~\citep{adam2026gptzerorobustdetectionllmgenerated}. Earlier detectors flagged essays by non-native English writers far more often than essays by native writers~\citep{liang2023gptdetectorsbiasednonnative}. OpenAI withdrew its own classifier in 2023 because of its low accuracy~\citep{openai2023classifier} before deploying textGrain in 2026.

\paragraph{AutoMark.} Perhaps most closely related to our work is that of
\citet{gloaguen2026automarkenablingautoresearchdiscover}.  In that work, the authors build a harness for the purpose of getting agents to propose new text watermarks, which are then automatically checked for distortion-free and soundness properties, before scoring detectability, robustness, and quality. While the objective of that work is very different from ours, in the process of developing the harness, they also study the soundness of existing watermarks. In particular, they, too, define soundness over the key (marginal validity) and soundness given a key (PAC-validity). They observe that marginal validity does not control the FPR for a fixed key and suggest using the same threshold as \cref{thm:markov}.  While these observations are used to inform their automated watermark proposer, they do not focus on evaluating existing watermarks. Our work instead focuses on evaluating existing watermarks, and we show that the conformalized approach used in SynthID fails under natural distribution shift.  We also provide a more complete empirical evaluation of the PAC-validity framework, showing that it can be applied to existing watermarks with only a small cost in TPR. Finally, conformalization appears in their work only as an option that agents may take, and they do not elaborate on the potential for distribution shift to invalidate a given guarantee.

\section{Discussion and Recommendations}\label{sec:discussion}

In this work, we have shown that existing watermarking schemes often fail to provide the advertised FPR control, and we have adopted a PAC-validity framework that can restore FPR control at a small cost in TPR.  We now discuss the implications of our findings and provide recommendations for the evaluation and design of future watermarking schemes.

We make three main recommendations for the evaluation and design of watermarking schemes going forward.  First, we recommend moving away from conformalized watermarks in high-stakes settings, where the cost of a false positive is high and the input distribution may shift in hard to predict ways.  Second, we recommend moving from marginal validity to PAC-validity, and adjusting the evidentiary thresholds for flagging content as AI-generated in order to do a better job of controlling the FPR for the deployed key.  Finally, we recommend that future watermarking schemes report the distribution of per-key FPRs over many keys and several datasets, performance on a single key, and the calibration data if used, in order to provide a more complete picture of the FPR control that can be expected in practice.

\paragraph{Limitations.}
While we believe our work provides a comprehensive evaluation of existing watermarking schemes, various limitations remain.  First, with at most $10{,}000$ samples per dataset, we have limited granularity in the estimation of the distribution of per-key FPRs, which is one of the reasons we defined keys as bad if they exceed $1.5\alpha$; arguments could be made for focusing on smaller thresholds, which would require testing on larger test sets.  Second, our null corpora are public datasets: code, chest X-rays, web screenshots, and the passage of time are examples of shifts a deployed detector could meet, but we do not have access to actual deployment data.  Third, while we mimicked a deployment implementation to the best of our ability, exact details of the deployed detectors are not public; moreover, we were forced to reimplement textGrain, since its detector is not public.  For SynthID, while we suspect that a version of the Bayesian detector from \citet{dathathri2024scalable} is used in deployment, as this is the primary version discussed in the paper, we cannot confirm this.

\bibliographystyle{plainnat}
\bibliography{references}

\begin{thebibliography}{65}
\providecommand{\natexlab}[1]{#1}
\providecommand{\url}[1]{\texttt{#1}}
\expandafter\ifx\csname urlstyle\endcsname\relax
  \providecommand{\doi}[1]{doi: #1}\else
  \providecommand{\doi}{doi: \begingroup \urlstyle{rm}\Url}\fi

\bibitem[Aaronson(2022)]{aaronson2022watermarking}
Scott Aaronson.
\newblock My {AI} safety lecture for {UT} effective altruism.
\newblock \url{https://scottaaronson.blog/?p=6823}, 2022.

\bibitem[Adam et~al.(2026)Adam, Cui, Thomas, Napier, Shmatko, Schnell, Tian, Dronavalli, Tian, and Lee]{adam2026gptzerorobustdetectionllmgenerated}
George~Alexandru Adam, Alexander Cui, Edwin Thomas, Emily Napier, Nazar Shmatko, Jacob Schnell, Jacob~Junqi Tian, Alekhya Dronavalli, Edward Tian, and Dongwon Lee.
\newblock Gptzero: Robust detection of llm-generated texts, 2026.
\newblock URL \url{https://arxiv.org/abs/2602.13042}.

\bibitem[Angelopoulos and Bates(2022)]{angelopoulos2022gentleintroductionconformalprediction}
Anastasios~N. Angelopoulos and Stephen Bates.
\newblock A gentle introduction to conformal prediction and distribution-free uncertainty quantification, 2022.
\newblock URL \url{https://arxiv.org/abs/2107.07511}.

\bibitem[{Anthropic}(2026)]{anthropic2026watermark}
{Anthropic}.
\newblock How {Claude}'s text watermarking works.
\newblock \url{https://www.anthropic.com/news/claude-text-watermark}, August 2026.
\newblock Published August 14, 2026.

\bibitem[Bates et~al.(2023)Bates, Candès, Lei, Romano, and Sesia]{Bates_2023}
Stephen Bates, Emmanuel Candès, Lihua Lei, Yaniv Romano, and Matteo Sesia.
\newblock Testing for outliers with conformal p-values.
\newblock \emph{The Annals of Statistics}, 51\penalty0 (1):\penalty0 149–178, February 2023.
\newblock ISSN 2168-8966.
\newblock \doi{10.1214/22-aos2244}.
\newblock URL \url{http://dx.doi.org/10.1214/22-AOS2244}.

\bibitem[Ben~Allal et~al.(2024)Ben~Allal, Lozhkov, Penedo, Wolf, and von Werra]{benallal2024cosmopedia}
Loubna Ben~Allal, Anton Lozhkov, Guilherme Penedo, Thomas Wolf, and Leandro von Werra.
\newblock Cosmopedia, 2024.
\newblock URL \url{https://huggingface.co/datasets/HuggingFaceTB/cosmopedia}.

\bibitem[Bleizeffer(2025)]{bleizeffer2025yale}
Kristy Bleizeffer.
\newblock Denying alleged {AI} use, {EMBA} student sues {Yale SOM} over year-long suspension.
\newblock Poets\&Quants, February 2025.
\newblock URL \url{https://poetsandquants.com/2025/02/25/denying-alleged-ai-use-emba-student-sues-yale-som-over-year-long-suspension/}.
\newblock February 25, 2025.

\bibitem[Block et~al.(2025)Block, Rakhlin, and Sekhari]{block2025gaussmark}
Adam Block, Alexander Rakhlin, and Ayush Sekhari.
\newblock {GaussMark}: A practical approach for structural watermarking of language models.
\newblock In \emph{Proceedings of the 42nd International Conference on Machine Learning}, volume 267 of \emph{Proceedings of Machine Learning Research}, pages 4507--4569. PMLR, 2025.
\newblock URL \url{https://proceedings.mlr.press/v267/block25a.html}.

\bibitem[Brown(2026)]{brown2026yangappeal}
Laura Brown.
\newblock {U of M} expulsion over alleged {AI} cheating upheld on appeal.
\newblock Minnesota Lawyer, February 2026.
\newblock URL \url{https://minnlawyer.com/2026/02/13/u-of-m-ai-cheating-expulsion-upheld-appeal/}.
\newblock February 13, 2026.

\bibitem[Chesney and Citron(2019)]{chesney2019deepfakes}
Robert~M. Chesney and Danielle~Keats Citron.
\newblock Deep fakes: A looming challenge for privacy, democracy, and national security.
\newblock \emph{California Law Review}, 107\penalty0 (6):\penalty0 1753--1820, 2019.

\bibitem[Christ et~al.(2024)Christ, Gunn, and Zamir]{pmlr-v247-christ24a}
Miranda Christ, Sam Gunn, and Or~Zamir.
\newblock Undetectable watermarks for language models.
\newblock In Shipra Agrawal and Aaron Roth, editors, \emph{Proceedings of Thirty Seventh Conference on Learning Theory}, volume 247 of \emph{Proceedings of Machine Learning Research}, pages 1125--1139. PMLR, 30 Jun--03 Jul 2024.
\newblock URL \url{https://proceedings.mlr.press/v247/christ24a.html}.

\bibitem[Cimpoi et~al.(2014)Cimpoi, Maji, Kokkinos, Mohamed, and Vedaldi]{cimpoi14describing}
M.~Cimpoi, S.~Maji, I.~Kokkinos, S.~Mohamed, and A.~Vedaldi.
\newblock Describing textures in the wild.
\newblock In \emph{Proceedings of the {IEEE} Conf. on Computer Vision and Pattern Recognition ({CVPR})}, 2014.

\bibitem[{CodeParrot}(2022)]{codeparrot2022clean}
{CodeParrot}.
\newblock {codeparrot-clean-valid}: deduplicated {Python} files from {GitHub}.
\newblock \url{https://huggingface.co/datasets/codeparrot/codeparrot-clean-valid}, 2022.

\bibitem[Cox et~al.(1997)Cox, Kilian, Leighton, and Shamoon]{650120}
I.J. Cox, J.~Kilian, F.T. Leighton, and T.~Shamoon.
\newblock Secure spread spectrum watermarking for multimedia.
\newblock \emph{IEEE Transactions on Image Processing}, 6\penalty0 (12):\penalty0 1673--1687, 1997.
\newblock \doi{10.1109/83.650120}.

\bibitem[Dathathri et~al.(2024)Dathathri, See, Ghaisas, Huang, McAdam, Welbl, Bachani, Kaskasoli, Stanforth, Matejovicova, et~al.]{dathathri2024scalable}
Sumanth Dathathri, Abigail See, Sumedh Ghaisas, Po-Sen Huang, Rob McAdam, Johannes Welbl, Vandana Bachani, Alex Kaskasoli, Robert Stanforth, Tatiana Matejovicova, et~al.
\newblock Scalable watermarking for identifying large language model outputs.
\newblock \emph{Nature}, 634\penalty0 (8035):\penalty0 818--823, 2024.

\bibitem[Diao et~al.(2025)Diao, Yang, Fu, Dong, Su, Kliegl, Chen, Belcak, Suhara, Yin, Patwary, Lin, Kautz, and Molchanov]{diao2025climb}
Shizhe Diao, Yu~Yang, Yonggan Fu, Xin Dong, Dan Su, Markus Kliegl, Zijia Chen, Peter Belcak, Yoshi Suhara, Hongxu Yin, Mostofa Patwary, Celine Lin, Jan Kautz, and Pavlo Molchanov.
\newblock Climb: Clustering-based iterative data mixture bootstrapping for language model pre-training.
\newblock \emph{arXiv preprint}, 2025.
\newblock URL \url{https://arxiv.org/abs/2504.13161}.

\bibitem[Emi and Spero(2024)]{emi2024technicalreportpangramaigenerated}
Bradley Emi and Max Spero.
\newblock Technical report on the pangram ai-generated text classifier, 2024.
\newblock URL \url{https://arxiv.org/abs/2402.14873}.

\bibitem[{European Commission}(2026)]{eucop2026}
{European Commission}.
\newblock Code of practice on marking and labelling of {AI}-generated content, 2026.
\newblock URL \url{https://digital-strategy.ec.europa.eu/en/policies/code-practice-ai-generated-content}.
\newblock Final version published 10 June 2026.

\bibitem[{European Parliament and Council of the European Union}(2024)]{euaiact2024}
{European Parliament and Council of the European Union}.
\newblock Regulation ({EU}) 2024/1689 laying down harmonised rules on artificial intelligence ({Artificial Intelligence Act}).
\newblock Official Journal of the European Union, L 2024/1689, 2024.
\newblock URL \url{https://eur-lex.europa.eu/eli/reg/2024/1689/oj}.
\newblock Article 50 applies from 2 August 2026.

\bibitem[Fernandez et~al.(2023{\natexlab{a}})Fernandez, Chaffin, Tit, Chappelier, and Furon]{fernandez2023bricksconsolidatewatermarkslarge}
Pierre Fernandez, Antoine Chaffin, Karim Tit, Vivien Chappelier, and Teddy Furon.
\newblock Three bricks to consolidate watermarks for large language models, 2023{\natexlab{a}}.
\newblock URL \url{https://arxiv.org/abs/2308.00113}.

\bibitem[Fernandez et~al.(2023{\natexlab{b}})Fernandez, Couairon, Jégou, Douze, and Furon]{fernandez2023stablesignaturerootingwatermarks}
Pierre Fernandez, Guillaume Couairon, Hervé Jégou, Matthijs Douze, and Teddy Furon.
\newblock The stable signature: Rooting watermarks in latent diffusion models, 2023{\natexlab{b}}.
\newblock URL \url{https://arxiv.org/abs/2303.15435}.

\bibitem[Fernandez et~al.(2024)Fernandez, Elsahar, Yalniz, and Mourachko]{fernandez2024videosealopenefficient}
Pierre Fernandez, Hady Elsahar, I.~Zeki Yalniz, and Alexandre Mourachko.
\newblock Video seal: Open and efficient video watermarking, 2024.
\newblock URL \url{https://arxiv.org/abs/2412.09492}.

\bibitem[Foundation()]{wikidump}
Wikimedia Foundation.
\newblock Wikimedia downloads.
\newblock URL \url{https://dumps.wikimedia.org}.

\bibitem[Gao et~al.(2020)Gao, Biderman, Black, Golding, Hoppe, Foster, Phang, He, Thite, Nabeshima, Presser, and Leahy]{gao2020pile800gbdatasetdiverse}
Leo Gao, Stella Biderman, Sid Black, Laurence Golding, Travis Hoppe, Charles Foster, Jason Phang, Horace He, Anish Thite, Noa Nabeshima, Shawn Presser, and Connor Leahy.
\newblock The pile: An 800gb dataset of diverse text for language modeling, 2020.
\newblock URL \url{https://arxiv.org/abs/2101.00027}.

\bibitem[Gerezgiher(2025)]{gerezgiher2025yang}
Feven Gerezgiher.
\newblock `{A} death penalty': {Ph.D.} student says {U of M} expelled him over unfair {AI} allegation.
\newblock MPR News, January 2025.
\newblock URL \url{https://www.mprnews.org/story/2025/01/17/phd-student-says-university-of-minnesota-expelled-him-over-ai-allegation}.
\newblock January 17, 2025.

\bibitem[Glickenhaus et~al.(2026)Glickenhaus, Thai, Russell, Masrour, Han, Spero, and Emi]{glickenhaus2026pangram4technicalreport}
Ben Glickenhaus, Katherine Thai, Jenna Russell, Elyas Masrour, Yue Han, Max Spero, and Bradley Emi.
\newblock Pangram 4 technical report, 2026.
\newblock URL \url{https://arxiv.org/abs/2607.27183}.

\bibitem[Gloaguen et~al.(2026)Gloaguen, Staab, and Vechev]{gloaguen2026automarkenablingautoresearchdiscover}
Thibaud Gloaguen, Robin Staab, and Martin Vechev.
\newblock Automark: Enabling autoresearch to discover better llm watermarks, 2026.
\newblock URL \url{https://arxiv.org/abs/2609.37310}.

\bibitem[Gowal et~al.(2025)Gowal, Bunel, Stimberg, Stutz, Ortiz-Jimenez, Kouridi, Vecerik, Hayes, Rebuffi, Bernard, Gamble, Horváth, Kaczmarczyck, Kaskasoli, Petrov, Shumailov, Thotakuri, Wiles, Yung, Ahmed, Martin, Rosen, Savčak, Senoner, Vyas, and Kohli]{gowal2025synthidimageimagewatermarkinginternet}
Sven Gowal, Rudy Bunel, Florian Stimberg, David Stutz, Guillermo Ortiz-Jimenez, Christina Kouridi, Mel Vecerik, Jamie Hayes, Sylvestre-Alvise Rebuffi, Paul Bernard, Chris Gamble, Miklós~Z. Horváth, Fabian Kaczmarczyck, Alex Kaskasoli, Aleksandar Petrov, Ilia Shumailov, Meghana Thotakuri, Olivia Wiles, Jessica Yung, Zahra Ahmed, Victor Martin, Simon Rosen, Christopher Savčak, Armin Senoner, Nidhi Vyas, and Pushmeet Kohli.
\newblock Synthid-image: Image watermarking at internet scale, 2025.
\newblock URL \url{https://arxiv.org/abs/2510.09263}.

\bibitem[{Gustavosta}(2022)]{gustavosta2022prompts}
{Gustavosta}.
\newblock Stable-diffusion-prompts.
\newblock \url{https://huggingface.co/datasets/Gustavosta/Stable-Diffusion-Prompts}, 2022.
\newblock Hugging Face dataset.

\bibitem[Hans et~al.(2024)Hans, Schwarzschild, Cherepanova, Kazemi, Saha, Goldblum, Geiping, and Goldstein]{hans2024spottingllmsbinocularszeroshot}
Abhimanyu Hans, Avi Schwarzschild, Valeriia Cherepanova, Hamid Kazemi, Aniruddha Saha, Micah Goldblum, Jonas Geiping, and Tom Goldstein.
\newblock Spotting llms with binoculars: Zero-shot detection of machine-generated text, 2024.
\newblock URL \url{https://arxiv.org/abs/2401.12070}.

\bibitem[Huang et~al.(2024)Huang, Zhu, Zhu, Ramchandran, Jordan, Lee, and Jiao]{huang2024optimalstatisticalwatermarking}
Baihe Huang, Hanlin Zhu, Banghua Zhu, Kannan Ramchandran, Michael~I. Jordan, Jason~D. Lee, and Jiantao Jiao.
\newblock Towards optimal statistical watermarking, 2024.
\newblock URL \url{https://arxiv.org/abs/2312.07930}.

\bibitem[Jovanović et~al.(2024)Jovanović, Staab, and Vechev]{jovanović2024watermarkstealinglargelanguage}
Nikola Jovanović, Robin Staab, and Martin Vechev.
\newblock Watermark stealing in large language models, 2024.
\newblock URL \url{https://arxiv.org/abs/2402.19361}.

\bibitem[Kermany(2018)]{kermany2018}
Daniel Kermany.
\newblock Labeled optical coherence tomography (oct) and chest x-ray images for classification, 2018.
\newblock URL \url{https://data.mendeley.com/datasets/rscbjbr9sj/2}.

\bibitem[Kirchenbauer et~al.(2024{\natexlab{a}})Kirchenbauer, Geiping, Wen, Katz, Miers, and Goldstein]{kirchenbauer2024watermarklargelanguagemodels}
John Kirchenbauer, Jonas Geiping, Yuxin Wen, Jonathan Katz, Ian Miers, and Tom Goldstein.
\newblock A watermark for large language models, 2024{\natexlab{a}}.
\newblock URL \url{https://arxiv.org/abs/2301.10226}.

\bibitem[Kirchenbauer et~al.(2024{\natexlab{b}})Kirchenbauer, Geiping, Wen, Shu, Saifullah, Kong, Fernando, Saha, Goldblum, and Goldstein]{kirchenbauer2024reliabilitywatermarkslargelanguage}
John Kirchenbauer, Jonas Geiping, Yuxin Wen, Manli Shu, Khalid Saifullah, Kezhi Kong, Kasun Fernando, Aniruddha Saha, Micah Goldblum, and Tom Goldstein.
\newblock On the reliability of watermarks for large language models, 2024{\natexlab{b}}.
\newblock URL \url{https://arxiv.org/abs/2306.04634}.

\bibitem[Kuditipudi et~al.(2024)Kuditipudi, Thickstun, Hashimoto, and Liang]{kuditipudi2024robustdistortionfreewatermarkslanguage}
Rohith Kuditipudi, John Thickstun, Tatsunori Hashimoto, and Percy Liang.
\newblock Robust distortion-free watermarks for language models, 2024.
\newblock URL \url{https://arxiv.org/abs/2307.15593}.

\bibitem[Laurençon et~al.(2024)Laurençon, Tronchon, and Sanh]{laurençon2024unlocking}
Hugo Laurençon, Léo Tronchon, and Victor Sanh.
\newblock Unlocking the conversion of web screenshots into html code with the websight dataset, 2024.

\bibitem[Lehmann and Romano(2022)]{lehmann2022testing}
E.L. Lehmann and Joseph~P. Romano.
\newblock \emph{Testing Statistical Hypotheses}.
\newblock Springer, Cham, 2022.
\newblock ISBN 978-3-030-70577-0.
\newblock \doi{10.1007/978-3-030-70578-7}.
\newblock URL \url{https://link.springer.com/book/10.1007/978-3-030-70578-7}.

\bibitem[Li et~al.(2025)Li, Ruan, Wang, Long, and Su]{Li_2025}
Xiang Li, Feng Ruan, Huiyuan Wang, Qi~Long, and Weijie~J. Su.
\newblock A statistical framework of watermarks for large language models: Pivot, detection efficiency and optimal rules.
\newblock \emph{The Annals of Statistics}, 53\penalty0 (1):\penalty0 322–351, February 2025.
\newblock ISSN 2168-8966.
\newblock \doi{10.1214/24-aos2468}.
\newblock URL \url{http://dx.doi.org/10.1214/24-AOS2468}.

\bibitem[Liang et~al.(2023)Liang, Yuksekgonul, Mao, Wu, and Zou]{liang2023gptdetectorsbiasednonnative}
Weixin Liang, Mert Yuksekgonul, Yining Mao, Eric Wu, and James Zou.
\newblock Gpt detectors are biased against non-native english writers, 2023.
\newblock URL \url{https://arxiv.org/abs/2304.02819}.

\bibitem[Lin et~al.(2015)Lin, Maire, Belongie, Bourdev, Girshick, Hays, Perona, Ramanan, Zitnick, and Dollár]{lin2015microsoftcococommonobjects}
Tsung-Yi Lin, Michael Maire, Serge Belongie, Lubomir Bourdev, Ross Girshick, James Hays, Pietro Perona, Deva Ramanan, C.~Lawrence Zitnick, and Piotr Dollár.
\newblock Microsoft coco: Common objects in context, 2015.
\newblock URL \url{https://arxiv.org/abs/1405.0312}.

\bibitem[Mitchell et~al.(2023)Mitchell, Lee, Khazatsky, Manning, and Finn]{mitchell2023detectgptzeroshotmachinegeneratedtext}
Eric Mitchell, Yoonho Lee, Alexander Khazatsky, Christopher~D. Manning, and Chelsea Finn.
\newblock Detectgpt: Zero-shot machine-generated text detection using probability curvature, 2023.
\newblock URL \url{https://arxiv.org/abs/2301.11305}.

\bibitem[Nilsback and Zisserman(2008)]{nilsback2008flowers}
Maria-Elena Nilsback and Andrew Zisserman.
\newblock Automated flower classification over a large number of classes.
\newblock In \emph{Indian Conference on Computer Vision, Graphics and Image Processing (ICVGIP)}, 2008.

\bibitem[Nilsson-Julien(2026)]{nilssonjulien2026netanyahu}
Estelle Nilsson-Julien.
\newblock Netanyahu denies death rumours in social media clip which {Grok} falsely branded {AI}.
\newblock Euronews, March 2026.
\newblock URL \url{https://www.euronews.com/my-europe/2026/03/18/netanyahu-denies-death-rumours-in-social-media-clip-which-grok-falsely-branded-ai}.

\bibitem[{OpenAI}(2023)]{openai2023classifier}
{OpenAI}.
\newblock New {AI} classifier for indicating {AI}-written text.
\newblock \url{https://openai.com/index/new-ai-classifier-for-indicating-ai-written-text/}, January 2023.
\newblock Discontinued on July 20, 2023.

\bibitem[{OpenAI}(2026{\natexlab{a}})]{openai2026textgrain}
{OpenAI}.
\newblock Our approach to {EU} text provenance rules.
\newblock \url{https://openai.com/index/eu-text-provenance/}, October 2026{\natexlab{a}}.

\bibitem[{OpenAI}(2026{\natexlab{b}})]{openai2026textgrainreport}
{OpenAI}.
\newblock {textGrain}: Entropy-calibrated watermarking for language model text.
\newblock Technical report, OpenAI, 2026{\natexlab{b}}.
\newblock URL \url{https://cdn.openai.com/pdf/e9508624-d767-41b6-a26d-e34ca798ada6/textgrain-entropy-calibrated-watermarking-for-language-model-text.pdf}.

\bibitem[Pang et~al.(2024)Pang, Hu, Zheng, and Smith]{pang2024freelunchllmwatermarking}
Qi~Pang, Shengyuan Hu, Wenting Zheng, and Virginia Smith.
\newblock No free lunch in llm watermarking: Trade-offs in watermarking design choices, 2024.
\newblock URL \url{https://arxiv.org/abs/2402.16187}.

\bibitem[Parkhi et~al.(2012)Parkhi, Vedaldi, Zisserman, and Jawahar]{parkhi2012pets}
Omkar~M. Parkhi, Andrea Vedaldi, Andrew Zisserman, and C.~V. Jawahar.
\newblock Cats and dogs.
\newblock In \emph{Proceedings of the IEEE Conference on Computer Vision and Pattern Recognition (CVPR)}, 2012.

\bibitem[Penedo et~al.(2024)Penedo, Kydl{\'\i}{\v{c}}ek, allal, Lozhkov, Mitchell, Raffel, Werra, and Wolf]{penedo2024the}
Guilherme Penedo, Hynek Kydl{\'\i}{\v{c}}ek, Loubna~Ben allal, Anton Lozhkov, Margaret Mitchell, Colin Raffel, Leandro~Von Werra, and Thomas Wolf.
\newblock The fineweb datasets: Decanting the web for the finest text data at scale.
\newblock In \emph{The Thirty-eight Conference on Neural Information Processing Systems Datasets and Benchmarks Track}, 2024.
\newblock URL \url{https://openreview.net/forum?id=n6SCkn2QaG}.

\bibitem[Piet et~al.(2024)Piet, Sitawarin, Fang, Mu, and Wagner]{piet2024markwordsanalyzingevaluating}
Julien Piet, Chawin Sitawarin, Vivian Fang, Norman Mu, and David Wagner.
\newblock Mark my words: Analyzing and evaluating language model watermarks, 2024.
\newblock URL \url{https://arxiv.org/abs/2312.00273}.

\bibitem[Raffel et~al.(2023)Raffel, Shazeer, Roberts, Lee, Narang, Matena, Zhou, Li, and Liu]{raffel2023exploringlimitstransferlearning}
Colin Raffel, Noam Shazeer, Adam Roberts, Katherine Lee, Sharan Narang, Michael Matena, Yanqi Zhou, Wei Li, and Peter~J. Liu.
\newblock Exploring the limits of transfer learning with a unified text-to-text transformer, 2023.
\newblock URL \url{https://arxiv.org/abs/1910.10683}.

\bibitem[Rombach et~al.(2022)Rombach, Blattmann, Lorenz, Esser, and Ommer]{rombach2022highresolutionimagesynthesislatent}
Robin Rombach, Andreas Blattmann, Dominik Lorenz, Patrick Esser, and Björn Ommer.
\newblock High-resolution image synthesis with latent diffusion models, 2022.
\newblock URL \url{https://arxiv.org/abs/2112.10752}.

\bibitem[Rose(2026)]{rose2026trust}
Adam Rose.
\newblock The real threat of {AI} is the collapse of trust.
\newblock Poynter (opinion), February 2026.
\newblock URL \url{https://www.poynter.org/commentary/2026/the-real-threat-of-ai-is-the-collapse-of-trust-deepfakes/}.

\bibitem[Sander et~al.(2026)Sander, Chang, Souček, Tran, Lacatusu, Rebuffi, Mourachko, Parimi, Ropers, Moritz, Stark, Elsahar, and Fernandez]{sander2026textseallocalizedllmwatermark}
Tom Sander, Hongyan Chang, Tomáš Souček, Tuan Tran, Valeriu Lacatusu, Sylvestre-Alvise Rebuffi, Alexandre Mourachko, Surya Parimi, Christophe Ropers, Rashel Moritz, Vanessa Stark, Hady Elsahar, and Pierre Fernandez.
\newblock Textseal: A localized llm watermark for provenance \& distillation protection, 2026.
\newblock URL \url{https://arxiv.org/abs/2605.12456}.

\bibitem[Shafer and Vovk(2007)]{shafer2007tutorialconformalprediction}
Glenn Shafer and Vladimir Vovk.
\newblock A tutorial on conformal prediction, 2007.
\newblock URL \url{https://arxiv.org/abs/0706.3188}.

\bibitem[Thomas et~al.(2026)Thomas, Block, and Goldblum]{thomas2026llmevaluationforecastingproblem}
Rahul Thomas, Adam Block, and Micah Goldblum.
\newblock Llm evaluation is a forecasting problem, 2026.
\newblock URL \url{https://arxiv.org/abs/2610.17018}.

\bibitem[{United States District Court for the District of Massachusetts}(2024)]{harris2024adams}
{United States District Court for the District of Massachusetts}.
\newblock Harris v. {Adams}, no. 24-cv-12437-pgl.
\newblock Memorandum and Order on Plaintiffs' Motion for Preliminary Injunction, November 2024.
\newblock URL \url{https://www.the74million.org/wp-content/uploads/2024/11/Hingham-judge-ruling-.pdf}.
\newblock Decided November 20, 2024.

\bibitem[Valiant(1984)]{Valiant1984ATO}
Leslie~G. Valiant.
\newblock A theory of the learnable.
\newblock \emph{Commun. ACM}, 27:\penalty0 1134--1142, 1984.
\newblock URL \url{https://api.semanticscholar.org/CorpusID:59712}.

\bibitem[Wen et~al.(2023)Wen, Kirchenbauer, Geiping, and Goldstein]{wen2023treeringwatermarksfingerprintsdiffusion}
Yuxin Wen, John Kirchenbauer, Jonas Geiping, and Tom Goldstein.
\newblock Tree-ring watermarks: Fingerprints for diffusion images that are invisible and robust, 2023.
\newblock URL \url{https://arxiv.org/abs/2305.20030}.

\bibitem[{Wikimedia Commons}(2026)]{wikimediacommons}
{Wikimedia Commons}.
\newblock Wikimedia {Commons}: a collection of freely usable media files.
\newblock \url{https://commons.wikimedia.org}, 2026.
\newblock Images uploaded 2013--2025, accessed November 2026.

\bibitem[Xie et~al.(2025)Xie, Chen, Ren, and Su]{xie2025watermarkclassroomconformalframework}
Yangxinyu Xie, Xuyang Chen, Zhimei Ren, and Weijie~J. Su.
\newblock Watermark in the classroom: A conformal framework for adaptive ai usage detection, 2025.
\newblock URL \url{https://arxiv.org/abs/2507.23113}.

\bibitem[Yang et~al.(2024)Yang, Zeng, Chen, Fang, Zhang, and Yu]{yang2024gaussianshadingprovableperformancelossless}
Zijin Yang, Kai Zeng, Kejiang Chen, Han Fang, Weiming Zhang, and Nenghai Yu.
\newblock Gaussian shading: Provable performance-lossless image watermarking for diffusion models, 2024.
\newblock URL \url{https://arxiv.org/abs/2404.04956}.

\bibitem[Zhao et~al.(2023)Zhao, Ananth, Li, and Wang]{zhao2023provablerobustwatermarkingaigenerated}
Xuandong Zhao, Prabhanjan Ananth, Lei Li, and Yu-Xiang Wang.
\newblock Provable robust watermarking for ai-generated text, 2023.
\newblock URL \url{https://arxiv.org/abs/2306.17439}.

\bibitem[Zhao et~al.(2026)Zhao, Wu, and Block]{zhao2026marktune}
Yizhou Zhao, Zhiwei~Steven Wu, and Adam Block.
\newblock {MarkTune}: Improving the quality-detectability trade-off in model-embedded {LLM} watermarking.
\newblock In \emph{Advances in Neural Information Processing Systems}, 2026.
\newblock URL \url{https://arxiv.org/abs/2512.04044}.
\newblock Forthcoming.

\end{thebibliography}

\appendix
\newpage

\section{Proofs}
\label{app:proofs}

This appendix proves the results of \cref{sec:pac_validity}. The key $\xi \sim \Pi$ and the content $X \sim P$ are independent, and $\FPR_P(\xi; t) = \pp_{X \sim P}(p_\xi(X) \le t)$ is the FPR of key $\xi$ when the detector is run at level $t$. We first show how PAC-validity relates to marginal validity (\cref{app:pac-marginal}), then prove \cref{thm:markov} (\cref{app:markov}), and finally show that this bound is tight (\cref{app:tight}).

\subsection{PAC-validity implies marginal validity}
\label{app:pac-marginal}

PAC-validity at target $\alpha$ is a stronger requirement than marginal validity at level $\alpha$, which controls not only the average FPR over keys but also how often a key is bad.

\begin{proposition}
\label{prop:pac-marginal}
If the detector run at level $t$ is $(\alpha', \delta)$-PAC valid under $P$, then the FPR averaged across all keys is bounded by $\ee_{\xi \sim \Pi}[\FPR_P(\xi; t)] \le \alpha' + \delta$.
\end{proposition}

\begin{proof}
A key is bad with probability at most $\delta$, in which case its FPR is at most $1$; otherwise its FPR is at most $\alpha'$. Hence
\[
  \ee_\xi\bigl[\FPR_P(\xi; t)\bigr] \;\le\; \alpha'\bigl(1 - \delta\bigr) + \delta
  \;\le\; \alpha' +\delta.
\]
This means $(\alpha', \delta)$-PAC-validity is sufficient for a target $\alpha$ such that $\alpha'+\delta \leq \alpha$.
\end{proof}

\subsection{Proof of \texorpdfstring{\cref{thm:markov}}{the Markov reduction}}
\label{app:markov}

If $\alpha' = 0$ the result is trivial.  Thus suppose $\alpha' > 0$.
The FPR of a key, $\FPR_P(\xi; \alpha'\delta)$, is a nonnegative random variable over $\xi \sim \Pi$. By Markov's inequality and marginal validity at level $\alpha'\delta$,
\[
  \pp_\xi\bigl(\FPR_P(\xi; \alpha'\delta) > \alpha'\bigr)
  \;\le\; \frac{\ee_\xi\bigl[\FPR_P(\xi; \alpha'\delta)\bigr]}{\alpha'}
  \;\le\; \frac{\alpha'\delta}{\alpha'} \;=\; \delta,
\]
so the detector run at level $\alpha'\delta$ is $(\alpha', \delta)$-PAC valid under $P$.

If the detector is pointwise valid, marginal validity holds under every $P$: since $\xi$ and $X$ are independent:
\[
  \ee_\xi\bigl[\FPR_P(\xi; t)\bigr] \;=\; \ee_{X \sim P}\bigl[\pp_\xi(p_\xi(X) \le t)\bigr] \;\leq\; t
\]
for every level $t$. The argument above then applies under every $P$. \qed

\subsection{The reduction from marginal to PAC-validity is tight}
\label{app:tight}

\Cref{thm:markov} runs the detector at threshold $\alpha'\delta$, which is much stricter than $\alpha'$. The next result shows that no larger threshold provides the same guarantee, even for a pointwise valid detector.

\begin{theorem}
\label{thm:tight}
For every $\alpha', \delta \in (0,1)$ and every level $t > \alpha'\delta$, there exist a key distribution $\Pi$, a content distribution $P$, and a pointwise valid detector such that the detector run at level $t$ is not $(\alpha', \delta)$-PAC valid under $P$.
\end{theorem}

The idea of the construction is to mix two detectors, chosen by the key: a bad detector, whose $p$-values are uniform on $[0, q]$, and a conservative one, whose $p$-values are uniform on $[q, 1]$. For any fixed piece of content, the mixture is exactly uniform, so the detector is pointwise valid. But each bad key flags too much content, and they occur more often than PAC-validity allows.

\begin{proof}
Since $t/\alpha' > \delta$, we can choose $q$ with $\delta < q < \min(1, t/\alpha')$. Let $P$ be the uniform distribution on $[0,1)$, and let the key $\xi = (\xi_1, \xi_2)$ be uniform on $[0,1)^2$. Let $w = (x + \xi_1) \bmod 1$, where $\bmod 1$  takes the fractional part. Define a detector such that a key is bad if $\xi_2 < q$:
\[
  p_\xi(x) =
  \begin{cases}
    q\,w & \text{if } \xi_2 < q, \\
    q + (1 - q)\,w & \text{otherwise.}
  \end{cases}
\]

\emph{The detector is pointwise valid.} For every fixed $x$, $w$ is uniform on $[0,1)$ because $\xi_1$ is. Hence, under a bad key, $p_\xi(x)$ is uniform on $[0, q)$, and under a good key, it is uniform on $[q, 1)$. Averaged over keys, $p_\xi(x)$ is uniform on $[0,1)$, so the detector is pointwise valid.

\emph{The detector is not PAC valid at level $t$.} Fix a bad key. Since $X$ is uniform on $[0,1)$, so is $(X + \xi_1) \bmod 1$, and therefore
\[
  \FPR_P(\xi; t) \;=\; \pp\bigl(q\,w \le t\bigr) \;>\; \alpha'.
\]
Bad keys occur with probability $q > \delta$, so $\pp_\xi(\FPR_P(\xi; t) > \alpha') \ge q > \delta$.
\end{proof}

\section{Where Bad Keys Come From}
\label{app:bad-keys}

A theoretical $p$-value makes a promise about a \emph{random} key: for any fixed passage, a randomly drawn key flags it with probability at most $\alpha$. A deployed watermark uses one \emph{fixed} key on many passages. If unwatermarked content has a common pattern, a fixed key either happens to line up with that pattern and flags too much, or does not, and flags too little. On average over keys, the two cancel, which is why the mean FPR is close to $\alpha$, but an individual key can be far off.

Most detectors we study score content $x$ with key $\xi$ via the inner product $T_\xi(x) = \langle a(\xi), b(x) \rangle$, where $a$ maps a key to a vector and $ b$ maps an input to a vector. For a fixed key, the average score over content from a distribution $P$ is $\langle a(\xi), \mu_P \rangle$, where $\mu_P = \ee_P[b(X)]$ is the average content vector. When $a(\xi)$ has mean $0$ over keys, this product is $0$ on average over keys but not for each key: a key whose vector points toward $\mu_P$ is more likely to detect false positives. Two datasets whose average vectors point the same way push the same keys up, so their bad keys coincide. What the common pattern $\mu_P$ is depends on the watermark.

\subsection{Text watermarks that hash the previous tokens}
\label{app:bad-keys-context}

KGW, Gumbel-max, TextSeal, SynthID-Base, and textGrain draw a pseudorandom value at each position using the key and the \emph{context}, the $L$ preceding tokens. KGW, Gumbel-max, and TextSeal use $L = 1$, while SynthID-Base and textGrain use $L = 4$. Here the common pattern is the set of (context, token) pairs that occur in many passages.

With $L = 1$, a few contexts such as ``the'', ``of'', and punctuation precede a large share of all text. If a key favors a common continuation of this context, for example, a KGW key whose green list after ``the'' contains common nouns, it raises the score of almost every input. Because these frequent pairs are shared by most English text, the same keys are bad on most datasets. With $L = 4$, there are vastly more contexts, and each occurs rarely, so no single context moves a key's scores much, and keys behave like the average key. The exception is text that repeats long phrases across documents, such as license headers and templates in code and Nemotron-CC.  SynthID-Base and textGrain have some bad keys, but only on individual datasets such as Code, rather than universally.

\begin{figure}[t]
    \centering
    \includegraphics[width=\linewidth]{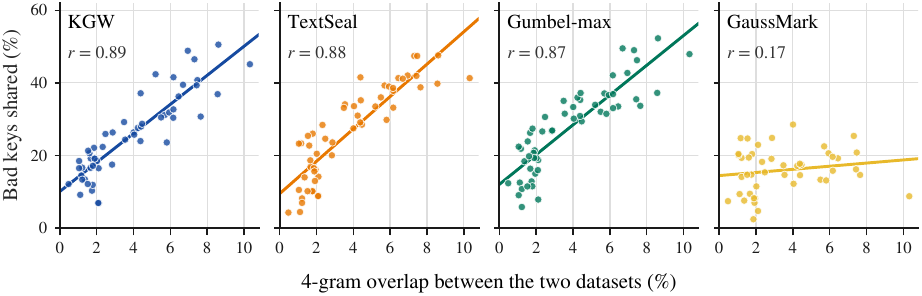}
    \caption{\textbf{Datasets that share more text share more of their bad keys.} Each dot is a pair of text datasets. Horizontal axis: share of one dataset's $4$-gram occurrences that also appear in the other. Vertical axis: share of the keys that are bad on one dataset (FPR $> 1.5\%$ at $\alpha = 1\%$) that are also bad on the other. SynthID-Base and textGrain have too few bad keys to plot.}
    \label{fig:badkey_scatter}
\end{figure}

\begin{figure}[t]
    \centering
    \includegraphics[width=0.75\linewidth]{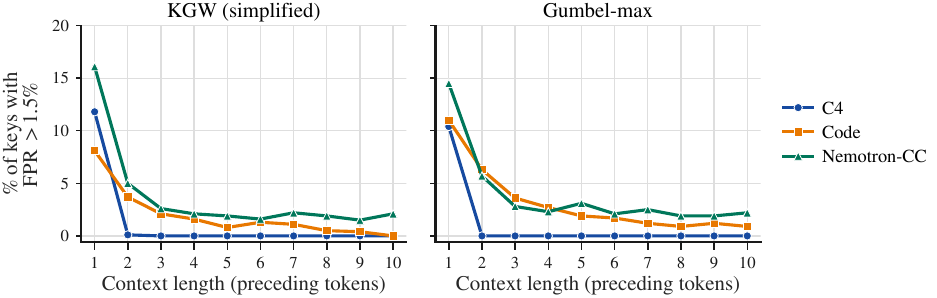}
    \caption{\textbf{Bad keys against context length.} Share of $1{,}000$ keys with FPR $> 1.5\%$ ($\alpha = 1\%$) for the simplified detectors of \cref{tab:context-length}, as the number of preceding tokens grows from $1$ to $10$.}
    \label{fig:context_sweep}
\end{figure}

Three results support this explanation. First, for KGW, TextSeal, and Gumbel-max, datasets that share more $4$-grams share more bad keys (\cref{fig:badkey_scatter}). Second, we changed the context length of KGW and Gumbel-max. With $L = 1$, they match the original schemes. With $L = 4$, the correlation across datasets drops, with almost no bad keys on C4, while some remain on code and Nemotron-CC (\cref{tab:context-length,fig:context_sweep}). The mean FPR is the same in both cases.

\begin{table}[t]
    \centering
    \caption{\textbf{A four-token context removes bad keys on ordinary text.} Share of $1{,}000$ keys with FPR $> 1.5\%$ ($\alpha = 1\%$) for simplified KGW and Gumbel-max detectors that use the $L$ preceding tokens. Last row: average correlation of per-key FPRs over the $55$ pairs of text datasets. KGW is a simplified version since the official KGW is infeasible to run beyond $L = 1$.}
    \label{tab:context-length}
    \small
    \begin{tabular}{@{}lcccc@{}}
        \toprule
        & \multicolumn{2}{c}{KGW (simplified)} & \multicolumn{2}{c}{Gumbel-max} \\
        \cmidrule(lr){2-3} \cmidrule(l){4-5}
        & $L = 1$ & $L = 4$ & $L = 1$ & $L = 4$ \\
        \midrule
        C4 (validation) & 11.8 & 0.0 & 10.4 & 0.0 \\
        Wikipedia & 10.8 & 0.0 & 13.8 & 0.0 \\
        Code & 8.1 & 1.6 & 11.0 & 2.7 \\
        Nemotron-CC & 16.1 & 2.1 & 14.5 & 2.3 \\
        \midrule
        Correlation across datasets & 0.34 & 0.00 & 0.33 & 0.01 \\
        \bottomrule
    \end{tabular}
\end{table}

\subsection{Watermarks without contexts}
\label{app:bad-keys-other}

\begin{figure}[t]
    \centering
    \includegraphics[width=\linewidth]{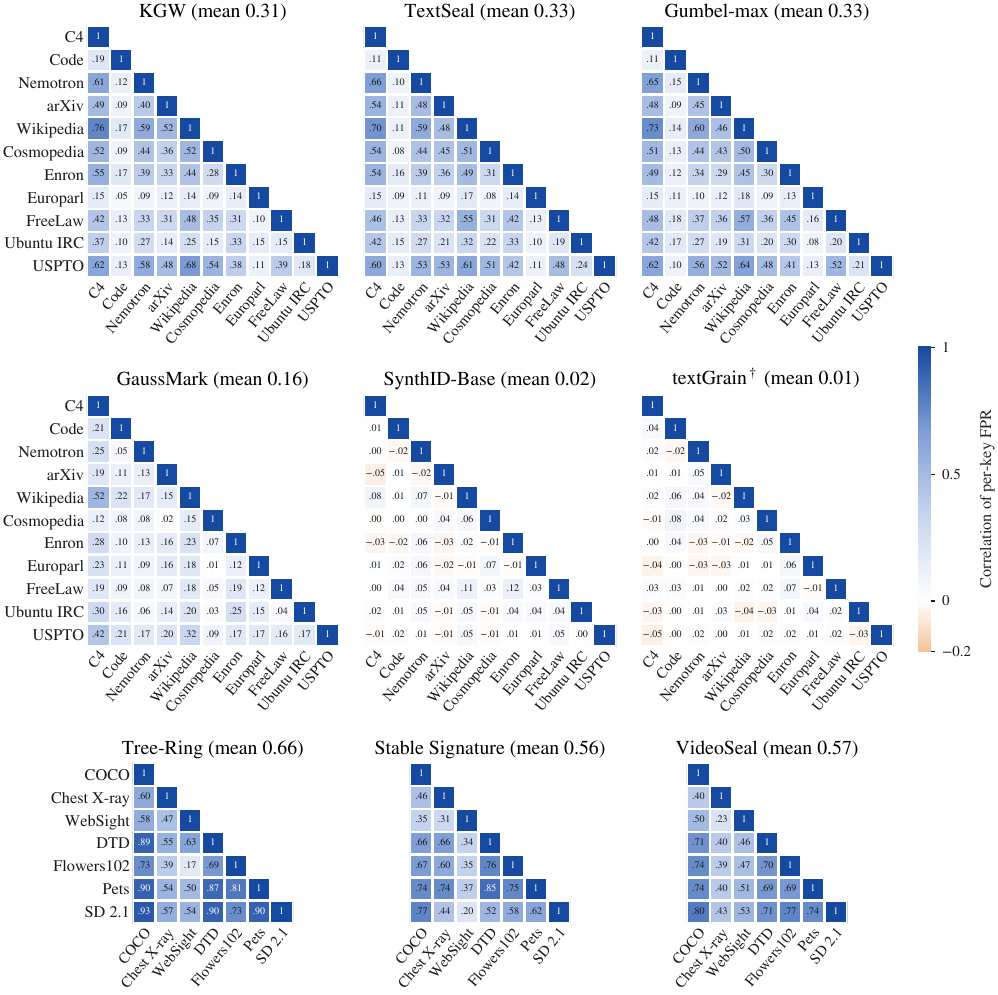}
\caption{\textbf{Cross-dataset correlation of per-key FPRs is strongest for the image watermarks.} Pairwise correlations, across $1{,}000$ keys, between the FPRs measured on different datasets at $\alpha=1\%$ under the theoretical cutoff. \textbf{Top:} KGW, TextSeal, and Gumbel-max. \textbf{Middle:} GaussMark, SynthID-Base, and textGrain$^\dagger$. \textbf{Bottom:} Tree-Ring, Stable Signature, and VideoSeal. $^\dagger$Our reimplementation.}
    \label{fig:fpr_correlation}
\end{figure}

\paragraph{GaussMark.}
GaussMark has no contexts or $n$-grams. Its key is a noise vector $Z$ that is added to the model weights, and its score is the normalized inner product of the gradient and $Z$. This score is exactly standard normal over the key \citep{block2025gaussmark}. For a fixed key, the common pattern is the average gradient direction of human-written text. This depends less on shared $n$-grams, which fits GaussMark's weaker correlations across datasets (\cref{fig:fpr_correlation}).

\paragraph{Image watermarks.}
Stable Signature and VideoSeal read a bit string from the image and count how many bits match the key. The bit string is extracted using an image decoder, and if the bit string is biased, a key that agrees with that bias matches more bits on every image. \citet{fernandez2023stablesignaturerootingwatermarks} report this bias for Stable Signature and correct it by whitening the extractor's output on a set of natural images. However, this bias is calculated using a calibration dataset, and suffers from the same exchangeability problem. This fits the much worse keys we see on chest X-rays (\Cref{fig:bars_conformal}) and WebSight screenshots than on COCO. For Tree-Ring, the common pattern would be shared structure in the Fourier coefficients recovered from images.

\subsection{Relation to prior work}
\label{app:bad-keys-related}

The context size used by the pseudorandom generator has been studied to determine robustness and security. Short contexts survive paraphrasing best \citep{kirchenbauer2024reliabilitywatermarkslargelanguage,piet2024markwordsanalyzingevaluating}, while long contexts are harder to steal or fake but easier to remove \citep{jovanović2024watermarkstealinglargelanguage,pang2024freelunchllmwatermarking}. The closest work on false positives is \citet{fernandez2023bricksconsolidatewatermarkslarge}. They found empirical FPRs above the promised ones and fixed this with exact tests and by scoring each repeated (context, token) pair once, but they measured the FPR pooled over $10$ keys. We tested larger context for KGW and Gumbel-max and found it reduces the probability of a bad key, but not completely. Other guarantees are also stated over a random key: Unigram-Watermark~\citep{zhao2023provablerobustwatermarkingaigenerated} uses one fixed green list (no context used in detection), and \citet{kuditipudi2024robustdistortionfreewatermarkslanguage} compute $p$-values against freshly drawn keys.

In short, longer contexts remove most bad keys on ordinary text, but they cost robustness and do not help on repetitive content, and watermarks without contexts have their own common patterns. A provider should therefore check the FPR of its own key on the content it expects, or use a cutoff that holds for most keys (\cref{sec:pac_validity}).

\section{Experimental Details}
\label{app:setup}

\subsection{Measuring the FPR of a key}
\label{app:setup-fpr}

For each watermark, we draw $1{,}000$ keys using random seeds. We compute the $ p$-value of each key across the dataset, and we call a key \emph{bad} if its FPR exceeds $1.5\alpha$. Unless stated otherwise, $\alpha = 1\%$. A text item is the first $1{,}000$ tokens of a document; we keep the first $10{,}000$ documents. Each image detector resizes images to its own input size, as in its official code ($512 \times 512$ for Tree-Ring and Stable Signature).

A perfectly conformalized detector also has bad keys, from sampling noise alone: with $n$ items, a key's number of flagged items is $\mathrm{Binomial}(n, \overline{\alpha})$, where $\overline{\alpha}$ is a random probability with mean at most $\alpha$ whose distribution is determined by the randomness in the selection of the content. For $n = 10{,}000$ and $\alpha = 1\%$, the probability that this exceeds $1.5\alpha n$ is below $0.1\%$, so a share of bad keys above that is not explained by noise.

\subsection{Datasets}
\label{app:setup-data}

\Cref{tab:datasets} lists the datasets. The calibration sets (C4 train and COCO) are disjoint from the held-out test sets. Code is Python from the CodeParrot dataset, and Nemotron-CC is topic cluster $16$ of the clustering of \citet{diao2025climb}. Chest X-rays are the training split of \citet{kermany2018}, and WebSight images are screenshots of rendered web pages. For the time series, we take one Common Crawl snapshot of FineWeb per year and, from Wikimedia Commons, camera photos taken and uploaded in each year; each key is conformalized on one year and tested on every later year, and results are averaged over the pairs of years with the same gap.

\subsection{True positive rates}
\label{app:setup-tpr}

\begin{figure}[t]
    \centering
    \includegraphics[width=\linewidth]{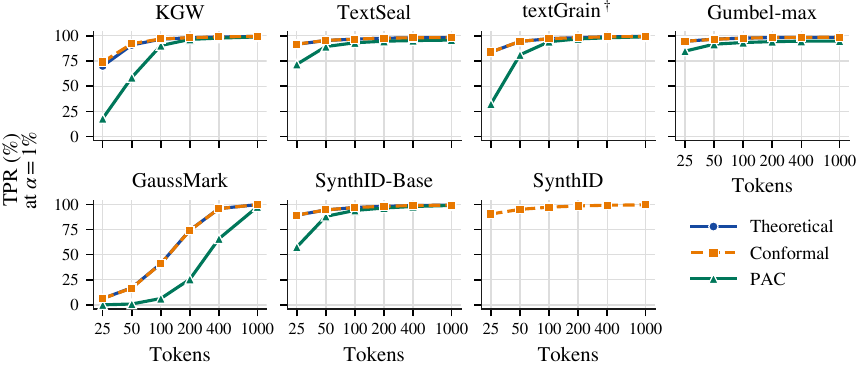}
\caption{\textbf{The same pattern holds when the text is generated by Llama-8B.} TPR against passage length for all seven text watermarks at target $\alpha=1\%$. Theoretical uses $p_\xi(x)\le\alpha$, Conformal uses per-key calibration, and PAC uses $p_\xi(x)\le\alpha^2/4$. SynthID has no theoretical $p$-value, so only its conformal curve is shown. $^\dagger$Our reimplementation.}
    \label{fig:tpr_llama8b}
\end{figure}

For each method, we generate watermarked content with the first $10$ of its $1{,}000$ keys (Stable Signature has a single released key). Text is generated from the first $30$ words of C4 validation text, $1{,}000$ tokens per text (temperature $1.0$, top-$p$ $0.95$). Images are scored as generated and after JPEG compression at quality $50$. Otherwise, generation and detection use the default settings and models of each official implementation (\cref{tab:setup_image,tab:setup_text}); SynthID's Bayesian detector is trained once and used for every key. In \cref{fig:tpr_llama8b}, every text scheme instead generates with Llama-3.1-8B, and GaussMark uses its paper's Llama setting (layer $28$ up-projection, variance $3 \times 10^{-4}$).

\begin{table}[t]
    \centering
    \caption{\textbf{Text watermark implementations and generation settings.} Generation models are used to evaluate true positive rates (TPRs). ``Model at detection'' indicates whether the detector requires the language model; otherwise, it uses the tokenizer. For each method, we use $10$ generation keys and generate $1{,}000$ texts per key, each $1{,}000$ tokens long, at temperature $1.0$, and top-$p$ $0.95$. Prompts are the first $30$ words of C4 validation passages.  $^\dagger$Our reimplementation.}
    \label{tab:setup_text}

    \footnotesize
    \setlength{\tabcolsep}{3pt}
    \renewcommand{\arraystretch}{1.15}
    \begin{tabular*}{\linewidth}{@{\extracolsep{\fill}}llcl@{}}
        \toprule
        \textbf{Method}
        & \textbf{Generation model}
        & \textbf{Model at detection?}
        & \textbf{Reference null} \\
        \midrule
        KGW~\citep{kirchenbauer2024watermarklargelanguagemodels}
        & OPT-1.3B & No & Binomial \\

        SynthID-Base~\citep{dathathri2024scalable}
        & Gemma-3-270M-IT & No & Binomial \\

        SynthID~\citep{dathathri2024scalable}
        & Gemma-3-270M-IT & No & -- (calibrated) \\

        TextSeal~\citep{sander2026textseallocalizedllmwatermark}
        & Llama-3.2-1B-Instruct & No & Gamma \\

        Gumbel-max~\citep{aaronson2022watermarking}
        & Llama-3.2-1B-Instruct & No & Gamma \\

        textGrain$^\dagger$~\citep{openai2026textgrainreport}
        & Llama-3.2-1B-Instruct & No & Gamma \\

        GaussMark~\citep{block2025gaussmark}
        & Phi-3.5-mini-Instruct & Yes & Normal \\
        \bottomrule
    \end{tabular*}
\end{table}

\begin{table}[t]
    \centering
    \caption{\textbf{Image watermark implementations and evaluation settings.}
    ``Model at detection'' indicates whether the detector requires the
    image generation model. Stable Signature and VideoSeal instead use
    their released extractor networks, with $48$-bit and $256$-bit
    signatures, respectively. All images are $512 \times 512$.
    For TPR evaluation, we use $10$ generation keys for each method,
    except Stable Signature, for which we use its one released key.
    We evaluate images both as generated and after JPEG compression
    at quality $50$.}
    \label{tab:setup_image}

    \footnotesize
    \setlength{\tabcolsep}{3pt}
    \renewcommand{\arraystretch}{1.15}
    \begin{tabular*}{\linewidth}{@{\extracolsep{\fill}}llcl@{}}
        \toprule
        \textbf{Method}
        & \textbf{Generation model}
        & \textbf{Model at detection?}
        & \textbf{Reference null} \\
        \midrule
        Tree-Ring~\citep{wen2023treeringwatermarksfingerprintsdiffusion}
        & SD~2.1 base
        & Yes & Non-central $\chi^2$ \\

        Stable Signature~\citep{fernandez2023stablesignaturerootingwatermarks}
        & SD~2.1 base (watermarked decoder)
        & No & Binomial \\

        VideoSeal~\citep{fernandez2024videosealopenefficient}
        & Post-hoc, on SD~2.1 base images
        & No & Binomial \\
        \bottomrule
    \end{tabular*}
\end{table}

\begin{table}[t]
    \centering
    \caption{\textbf{Datasets and evaluation roles.}
    Text inputs are the first $1{,}000$ tokens of each document.
    ``Primary'' identifies the principal evaluation datasets;
    ``additional'' identifies the expanded evaluation set.
    For the time-series evaluation, each key is calibrated on one year's
    data and tested on every later year. Time-series sample counts are
    per year.}
    \label{tab:datasets}

    \footnotesize
    \setlength{\tabcolsep}{4pt}
    \renewcommand{\arraystretch}{1.1}
    \begin{tabular*}{\linewidth}{@{\extracolsep{\fill}}llr@{}}
        \toprule
        \textbf{Dataset} & \textbf{Role} & \textbf{Size} \\
        \midrule
        \multicolumn{3}{@{}l}{\textit{Text}} \\

        C4 train~\citep{raffel2023exploringlimitstransferlearning}
        & Calibration (primary) & $10{,}000$ \\

        C4 validation~\citep{raffel2023exploringlimitstransferlearning}
        & Held-out test (primary) & $10{,}000$ \\

        Code (Python)~\citep{codeparrot2022clean}
        & Shift (primary) & $10{,}000$ \\

        Nemotron-CC, cluster 16~\citep{diao2025climb}
        & Shift (primary) & $10{,}000$ \\

        FineWeb, 2013--2025~\citep{penedo2024the}
        & Time series (primary) & $10{,}000$ per year \\

        arXiv~\citep{gao2020pile800gbdatasetdiverse}
        & Shift (additional) & $10{,}000$ \\

        Europarl~\citep{gao2020pile800gbdatasetdiverse}
        & Shift (additional) & $10{,}000$ \\

        FreeLaw~\citep{gao2020pile800gbdatasetdiverse}
        & Shift (additional) & $10{,}000$ \\

        Enron emails~\citep{gao2020pile800gbdatasetdiverse}
        & Shift (additional) & $10{,}000$ \\

        Ubuntu IRC~\citep{gao2020pile800gbdatasetdiverse}
        & Shift (additional) & $10{,}000$ \\

        USPTO~\citep{gao2020pile800gbdatasetdiverse}
        & Shift (additional) & $10{,}000$ \\

        Wikipedia (English)~\citep{wikidump}
        & Shift (additional) & $10{,}000$ \\

        Cosmopedia~\citep{benallal2024cosmopedia}
        & Shift (additional) & $10{,}000$ \\
        \midrule

        \multicolumn{3}{@{}l}{\textit{Images}} \\

        COCO (val2017 + 5k test2017)~\citep{lin2015microsoftcococommonobjects}
        & Calibration (primary) & $10{,}000$ \\

        COCO (held out, test2017)~\citep{lin2015microsoftcococommonobjects}
        & Held-out test (primary) & $10{,}000$ \\

        Chest X-ray~\citep{kermany2018}
        & Shift (primary) & $5{,}216$ \\

        WebSight~\citep{laurençon2024unlocking}
        & Shift (primary) & $10{,}000$ \\

        Wikimedia Commons, 2013--2025~\citep{wikimediacommons}
        & Time series (primary) & $10{,}000$ per year \\

        DTD~\citep{cimpoi14describing}
        & Shift (additional) & $5{,}640$ \\

        Flowers102~\citep{nilsback2008flowers}
        & Shift (additional) & $8{,}189$ \\

        Oxford-IIIT Pets~\citep{parkhi2012pets}
        & Shift (additional) & $7{,}390$ \\

        SD~2.1, generated~\citep{rombach2022highresolutionimagesynthesislatent,gustavosta2022prompts}
        & Shift (additional) & $10{,}000$ \\
        \bottomrule
    \end{tabular*}
\end{table}

\section{Additional Results}
\label{app:additional-results}

\begin{figure}[t]
    \centering
    \includegraphics[width=\linewidth]{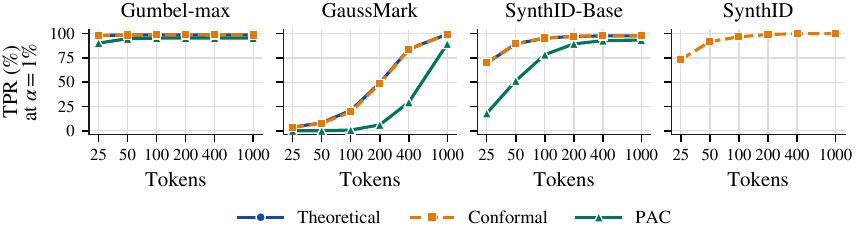}
\caption{\textbf{The power cost of the PAC cutoff varies substantially across watermarks.} TPR against passage length for Gumbel-max, GaussMark, SynthID-Base, and SynthID. At target $\alpha=1\%$, Theoretical uses $p_\xi(x)\le\alpha$, Conformal uses per-key calibration on C4 training data, and PAC uses $p_\xi(x)\le\alpha^2/4$. SynthID has no theoretical $p$-value, so only its conformal curve is shown.}
    \label{fig:tpr_length_appendix}
\end{figure}

\begin{figure}[t]
    \centering
    \includegraphics[width=\linewidth]{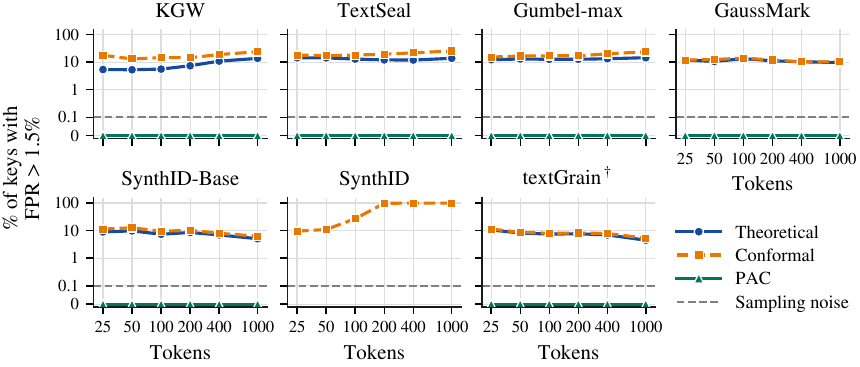}
\caption{\textbf{Bad keys occur at every passage length under the theoretical and conformal cutoffs; none are observed under the PAC cutoff.} Highest percentage of $1{,}000$ keys across C4 validation, Code, and Nemotron-CC whose FPR exceeds $1.5\%$ at target $\alpha=1\%$, against passage length. Theoretical uses $p_\xi(x)\le\alpha$, Conformal uses per-key calibration, and PAC uses $p_\xi(x)\le\alpha^2/4$. SynthID has no theoretical $p$-value, so only its conformal curve is shown. Dashed lines mark the sampling-noise reference.  $^\dagger$Our reimplementation.}
    \label{fig:badkeys_length}
\end{figure}

\begin{table*}[t]
    \centering
    \caption{\textbf{FPRs of text watermarks under theoretical cutoffs.}
    Results are for $1{,}000$-token passages using $p_\xi(x) \le \alpha$.
    \emph{Mean} is the FPR averaged over $1{,}000$ keys;
    \emph{$>1.5\alpha$} is the percentage of keys whose false positive
    rate exceeds $1.5\alpha$. $^\dagger$Our reimplementation.}
    \label{tab:marginal_text}

    \scriptsize
    \setlength{\tabcolsep}{2pt}
    \renewcommand{\arraystretch}{1.1}
    \begin{tabular*}{\linewidth}{@{\extracolsep{\fill}}l*{12}{r}@{}}
        \toprule
        & \multicolumn{2}{c}{\textbf{KGW}~\citep{kirchenbauer2024watermarklargelanguagemodels}}
        & \multicolumn{2}{c}{\textbf{TextSeal}~\citep{sander2026textseallocalizedllmwatermark}}
        & \multicolumn{2}{c}{\textbf{Gumbel-max}~\citep{aaronson2022watermarking}}
        & \multicolumn{2}{c}{\textbf{GaussMark}~\citep{block2025gaussmark}}
        & \multicolumn{2}{c}{\textbf{SynthID-Base}~\citep{dathathri2024scalable}}
        & \multicolumn{2}{c}{\textbf{textGrain}$^\dagger$~\citep{openai2026textgrainreport}} \\
        \cmidrule(lr){2-3}
        \cmidrule(lr){4-5}
        \cmidrule(lr){6-7}
        \cmidrule(lr){8-9}
        \cmidrule(lr){10-11}
        \cmidrule(lr){12-13}
        \textbf{Dataset}
        & Mean & $>1.5\alpha$
        & Mean & $>1.5\alpha$
        & Mean & $>1.5\alpha$
        & Mean & $>1.5\alpha$
        & Mean & $>1.5\alpha$
        & Mean & $>1.5\alpha$ \\
        \midrule

        \multicolumn{13}{@{}l}{\textbf{\boldmath $\alpha = 1\%$}} \\
        C4 (validation) & 0.89 & 9.6 & 0.87 & 9.1 & 0.93 & 10.4 & 1.00 & 2.4 & 0.98 & 0.0 & 1.00 & 0.0 \\
        Code & 0.85 & 7.5 & 0.89 & 10.9 & 0.90 & 11.0 & 1.00 & 2.5 & 0.97 & 3.8 & 1.00 & 4.4 \\
        arXiv & 0.87 & 11.6 & 0.89 & 13.4 & 0.93 & 13.9 & 0.99 & 14.6 & 0.98 & 0.8 & 1.00 & 0.2 \\
        Wikipedia & 0.87 & 9.4 & 0.87 & 9.4 & 0.95 & 13.8 & 1.00 & 0.4 & 0.99 & 0.0 & 1.00 & 0.0 \\
        Cosmopedia & 0.89 & 10.9 & 0.86 & 9.3 & 0.92 & 10.5 & 0.97 & 15.8 & 0.99 & 0.0 & 1.00 & 0.0 \\
        Enron emails & 0.89 & 7.9 & 0.88 & 7.8 & 0.94 & 11.2 & 0.98 & 12.0 & 0.98 & 2.1 & 0.99 & 1.4 \\
        Europarl & 0.86 & 5.9 & 0.89 & 3.0 & 0.93 & 4.6 & 0.98 & 10.8 & 0.98 & 0.0 & 1.00 & 0.0 \\
        FreeLaw & 0.84 & 11.4 & 0.89 & \textbf{14.2} & 0.94 & \textbf{14.7} & 1.00 & 14.1 & 0.99 & 0.6 & 1.00 & 0.5 \\
        Ubuntu IRC & 0.90 & 11.9 & 0.90 & 11.6 & 0.94 & 11.9 & 0.99 & \textbf{17.1} & 0.98 & 0.4 & 1.00 & 0.0 \\
        USPTO & 0.89 & \textbf{14.7} & 0.89 & 13.8 & 0.93 & 14.3 & 1.00 & 6.8 & 0.99 & 0.3 & 1.00 & 0.7 \\
        Nemotron-CC & 0.88 & 13.7 & 0.87 & 13.8 & 0.92 & 14.5 & 1.00 & 9.7 & 0.98 & \textbf{5.1} & 1.00 & \textbf{4.5} \\
        \midrule

        \multicolumn{13}{@{}l}{\textbf{\boldmath $\alpha = 5\%$}} \\
        C4 (validation) & 4.56 & 6.4 & 4.53 & 6.3 & 4.73 & 7.7 & 5.00 & 0.5 & 4.94 & 0.0 & 5.00 & 0.0 \\
        Code & 4.42 & 4.6 & 4.60 & 8.6 & 4.63 & 8.6 & 5.00 & 0.4 & 4.92 & \textbf{1.2} & 4.99 & \textbf{1.1} \\
        arXiv & 4.50 & 9.6 & 4.61 & 10.8 & 4.74 & 11.3 & 4.97 & 11.8 & 4.92 & 0.1 & 4.99 & 0.0 \\
        Wikipedia & 4.53 & 5.7 & 4.55 & 6.7 & 4.80 & 10.2 & 5.01 & 0.2 & 4.94 & 0.0 & 4.99 & 0.0 \\
        Cosmopedia & 4.57 & 7.3 & 4.52 & 7.1 & 4.70 & 7.8 & 4.88 & \textbf{14.0} & 4.92 & 0.0 & 5.01 & 0.0 \\
        Enron emails & 4.54 & 4.4 & 4.59 & 6.8 & 4.79 & 8.3 & 4.97 & 7.5 & 4.95 & 0.1 & 4.98 & 0.0 \\
        Europarl & 4.46 & 2.3 & 4.61 & 0.4 & 4.75 & 1.6 & 4.94 & 6.9 & 4.91 & 0.0 & 4.99 & 0.0 \\
        FreeLaw & 4.37 & 8.9 & 4.64 & 11.1 & 4.79 & \textbf{12.0} & 5.01 & 9.0 & 4.95 & 0.0 & 5.01 & 0.0 \\
        Ubuntu IRC & 4.59 & 8.8 & 4.66 & 8.1 & 4.80 & 8.5 & 4.97 & 13.3 & 4.92 & 0.0 & 5.02 & 0.0 \\
        USPTO & 4.55 & \textbf{11.9} & 4.59 & \textbf{11.3} & 4.75 & 11.9 & 5.00 & 3.5 & 4.96 & 0.0 & 5.01 & 0.1 \\
        Nemotron-CC & 4.58 & 11.0 & 4.53 & 10.8 & 4.74 & 11.1 & 5.01 & 3.3 & 4.95 & 0.1 & 4.99 & 0.0 \\
        \bottomrule
    \end{tabular*}
\end{table*}

\begin{table*}[t]
    \centering
    \caption{\textbf{FPRs of image watermarks under theoretical cutoffs.}
    Detectors use the cutoff $p_\xi(x) \le \alpha$.
    \emph{Mean} is the FPR averaged over $1{,}000$ keys;
    \emph{$>1.5\alpha$} is the percentage of keys whose false positive
    rate exceeds $1.5\alpha$. }
    \label{tab:marginal_image}

    \footnotesize
    \setlength{\tabcolsep}{3pt}
    \renewcommand{\arraystretch}{1.1}
    \begin{tabular*}{\linewidth}{@{\extracolsep{\fill}}l*{6}{r}@{}}
        \toprule
        & \multicolumn{2}{c}{\textbf{Tree-Ring}~\citep{wen2023treeringwatermarksfingerprintsdiffusion}}
        & \multicolumn{2}{c}{\textbf{Stable Signature}~\citep{fernandez2023stablesignaturerootingwatermarks}}
        & \multicolumn{2}{c}{\textbf{VideoSeal}~\citep{fernandez2024videosealopenefficient}} \\
        \cmidrule(lr){2-3}
        \cmidrule(lr){4-5}
        \cmidrule(lr){6-7}
        \textbf{Dataset}
        & Mean & $>1.5\alpha$
        & Mean & $>1.5\alpha$
        & Mean & $>1.5\alpha$ \\
        \midrule

        \multicolumn{7}{@{}l}{\textbf{\boldmath $\alpha = 1\%$}} \\
        COCO (held out) & 0.29 & 0.0 & 0.66 & 8.0 & 0.73 & 0.4 \\
        SD~2.1 (generated) & 0.29 & 0.0 & 0.70 & \textbf{11.9} & 0.73 & 1.1 \\
        DTD & 0.33 & 0.5 & 0.67 & 8.2 & 0.72 & 0.2 \\
        Flowers102 & 0.25 & 0.4 & 0.67 & 11.8 & 0.72 & 0.2 \\
        Oxford-IIIT Pets & 0.27 & 0.0 & 0.67 & 9.7 & 0.73 & 0.1 \\
        WebSight & 0.45 & \textbf{4.8} & 0.61 & 9.6 & 0.72 & 0.5 \\
        Chest X-ray & 0.48 & 4.3 & 0.70 & 9.4 & 0.72 & \textbf{4.3} \\
        \midrule

        \multicolumn{7}{@{}l}{\textbf{\boldmath $\alpha = 5\%$}} \\
        COCO (held out) & 2.28 & 0.0 & 2.95 & 3.7 & 4.55 & 0.2 \\
        SD~2.1 (generated) & 2.30 & 0.0 & 3.09 & 9.0 & 4.53 & 1.4 \\
        DTD & 2.42 & 0.0 & 3.01 & 4.3 & 4.53 & 0.2 \\
        Flowers102 & 2.13 & 0.3 & 3.01 & 10.2 & 4.53 & 0.1 \\
        Oxford-IIIT Pets & 2.19 & 0.0 & 3.00 & 8.0 & 4.54 & 0.0 \\
        WebSight & 2.62 & 2.8 & 2.84 & 9.9 & 4.52 & 0.6 \\
        Chest X-ray & 3.08 & \textbf{4.1} & 3.12 & \textbf{11.3} & 4.53 & \textbf{5.1} \\
        \bottomrule
    \end{tabular*}
\end{table*}

\begin{table*}[t]
    \centering
    \caption{\textbf{FPR of text watermarks after per-key calibration.}
    Each key's conformal cutoff is calibrated at target level $\alpha$
    on $10{,}000$ C4 training passages. Text passages contain $1{,}000$ tokens.
    \emph{Mean} is the FPR averaged over $1{,}000$ keys;
    \emph{$>1.5\alpha$} is the percentage of keys whose false positive
    rate exceeds $1.5\alpha$. $^\dagger$Our reimplementation.}
    \label{tab:conformal_text}

    \scriptsize
    \setlength{\tabcolsep}{1.5pt}
    \renewcommand{\arraystretch}{1.1}
    \resizebox{\linewidth}{!}{%
    \begin{tabular}{@{}l*{14}{r}@{}}
        \toprule
        & \multicolumn{2}{c}{\textbf{KGW}~\citep{kirchenbauer2024watermarklargelanguagemodels}}
        & \multicolumn{2}{c}{\textbf{TextSeal}~\citep{sander2026textseallocalizedllmwatermark}}
        & \multicolumn{2}{c}{\textbf{Gumbel-max}~\citep{aaronson2022watermarking}}
        & \multicolumn{2}{c}{\textbf{GaussMark}~\citep{block2025gaussmark}}
        & \multicolumn{2}{c}{\textbf{SynthID-Base}~\citep{dathathri2024scalable}}
        & \multicolumn{2}{c}{\textbf{SynthID}~\citep{dathathri2024scalable}}
        & \multicolumn{2}{c}{\textbf{textGrain}$^\dagger$~\citep{openai2026textgrainreport}} \\
        \cmidrule(lr){2-3}
        \cmidrule(lr){4-5}
        \cmidrule(lr){6-7}
        \cmidrule(lr){8-9}
        \cmidrule(lr){10-11}
        \cmidrule(lr){12-13}
        \cmidrule(lr){14-15}
        \textbf{Dataset}
        & Mean & $>1.5\alpha$
        & Mean & $>1.5\alpha$
        & Mean & $>1.5\alpha$
        & Mean & $>1.5\alpha$
        & Mean & $>1.5\alpha$
        & Mean & $>1.5\alpha$
        & Mean & $>1.5\alpha$ \\
        \midrule

        \multicolumn{15}{@{}l}{\textbf{\boldmath $\alpha = 1\%$}} \\
        C4 (validation) & 1.00 & 0.1 & 1.01 & 0.0 & 1.01 & 0.2 & 1.00 & 0.0 & 1.00 & 0.0 & 1.12 & 0.0 & 1.00 & 0.0 \\
        Code & 1.13 & \textbf{23.8} & 1.21 & 25.3 & 1.15 & 23.8 & 1.02 & 7.4 & 0.99 & 5.1 & 34.12 & \textbf{100.0} & 1.00 & 4.6 \\
        arXiv & 1.05 & 16.8 & 1.08 & 19.9 & 1.07 & 19.2 & 1.00 & \textbf{16.7} & 1.00 & 2.1 & 2.04 & 95.3 & 1.00 & 0.5 \\
        Wikipedia & 1.03 & 8.5 & 1.05 & 12.6 & 1.05 & 11.5 & 1.02 & 1.6 & 1.00 & 0.1 & 0.40 & 0.0 & 1.00 & 0.2 \\
        Cosmopedia & 1.09 & 19.3 & 1.08 & 17.4 & 1.08 & 17.8 & 0.99 & 16.5 & 1.00 & 0.1 & 0.33 & 0.0 & 1.00 & 0.2 \\
        Enron emails & 1.10 & 18.3 & 1.11 & 19.5 & 1.12 & 19.3 & 0.98 & 12.4 & 1.00 & 3.7 & 10.69 & \textbf{100.0} & 0.99 & 1.6 \\
        Europarl & 1.15 & 22.9 & 1.20 & \textbf{25.6} & 1.19 & \textbf{25.9} & 0.99 & 11.4 & 0.99 & 0.3 & 20.41 & \textbf{100.0} & 1.00 & 0.2 \\
        FreeLaw & 1.03 & 17.8 & 1.11 & 21.6 & 1.08 & 19.9 & 1.02 & 14.9 & 1.00 & 2.7 & 20.76 & \textbf{100.0} & 1.00 & 0.8 \\
        Ubuntu IRC & 1.13 & 23.6 & 1.15 & 22.3 & 1.14 & 21.8 & 0.99 & 16.4 & 1.00 & 1.3 & 7.65 & \textbf{100.0} & 1.00 & 0.6 \\
        USPTO & 1.01 & 15.6 & 1.04 & 14.7 & 1.02 & 13.0 & 1.00 & 6.2 & 1.01 & 1.6 & 1.29 & 12.6 & 0.99 & 0.9 \\
        Nemotron-CC & 1.01 & 15.2 & 1.02 & 14.3 & 1.02 & 15.5 & 1.01 & 10.2 & 1.00 & \textbf{6.3} & 0.43 & 0.1 & 0.99 & \textbf{5.3} \\
        \midrule

        \multicolumn{15}{@{}l}{\textbf{\boldmath $\alpha = 5\%$}} \\
        C4 (validation) & 5.00 & 0.0 & 5.01 & 0.0 & 5.01 & 0.0 & 5.00 & 0.0 & 5.00 & 0.0 & 5.11 & 0.0 & 4.99 & 0.0 \\
        Code & 5.31 & \textbf{18.2} & 5.55 & \textbf{20.3} & 5.37 & \textbf{18.0} & 5.07 & 1.9 & 4.99 & \textbf{1.0} & 70.01 & \textbf{100.0} & 4.98 & \textbf{1.2} \\
        arXiv & 5.14 & 12.4 & 5.25 & 14.1 & 5.17 & 13.7 & 5.00 & 12.5 & 4.99 & 0.1 & 12.99 & \textbf{100.0} & 4.99 & 0.0 \\
        Wikipedia & 5.09 & 3.4 & 5.14 & 6.5 & 5.17 & 6.1 & 5.06 & 0.0 & 5.01 & 0.0 & 3.73 & 0.0 & 4.99 & 0.0 \\
        Cosmopedia & 5.24 & 13.0 & 5.23 & 12.3 & 5.19 & 11.8 & 4.93 & \textbf{13.9} & 4.99 & 0.0 & 3.96 & 0.0 & 5.01 & 0.0 \\
        Enron emails & 5.24 & 10.3 & 5.33 & 10.7 & 5.33 & 12.4 & 4.99 & 7.4 & 5.02 & 0.2 & 26.26 & \textbf{100.0} & 4.98 & 0.0 \\
        Europarl & 5.38 & 16.2 & 5.57 & 16.2 & 5.51 & 16.4 & 4.98 & 6.4 & 4.98 & 0.0 & 54.74 & \textbf{100.0} & 4.99 & 0.0 \\
        FreeLaw & 5.02 & 13.3 & 5.33 & 16.9 & 5.25 & 14.0 & 5.06 & 9.8 & 5.02 & 0.1 & 31.93 & \textbf{100.0} & 5.00 & 0.0 \\
        Ubuntu IRC & 5.35 & 16.9 & 5.44 & 15.2 & 5.39 & 14.6 & 4.98 & 11.7 & 4.99 & 0.0 & 29.59 & \textbf{100.0} & 5.01 & 0.0 \\
        USPTO & 5.04 & 10.7 & 5.13 & 10.9 & 5.06 & 9.0 & 5.01 & 2.3 & 5.03 & 0.1 & 8.06 & 76.7 & 5.01 & 0.0 \\
        Nemotron-CC & 5.07 & 10.0 & 5.06 & 8.4 & 5.08 & 8.9 & 5.05 & 3.5 & 5.01 & 0.1 & 4.56 & 0.0 & 4.99 & 0.0 \\
        \bottomrule
    \end{tabular}%
    }
\end{table*}

\begin{table*}[t]
    \centering
    \caption{\textbf{False-positive rates of image watermarks after per-key calibration.}
    Each key's conformal cutoff is calibrated at target level $\alpha$
    on $10{,}000$ COCO images.
    \emph{Mean} is the false positive rate averaged over $1{,}000$ keys;
    \emph{$>1.5\alpha$} is the percentage of keys whose false positive
    rate exceeds $1.5\alpha$. }
    \label{tab:conformal_image}

    \footnotesize
    \setlength{\tabcolsep}{3pt}
    \renewcommand{\arraystretch}{1.1}
    \begin{tabular*}{\linewidth}{@{\extracolsep{\fill}}l*{6}{r}@{}}
        \toprule
        & \multicolumn{2}{c}{\textbf{Tree-Ring}~\citep{wen2023treeringwatermarksfingerprintsdiffusion}}
        & \multicolumn{2}{c}{\textbf{Stable Signature}~\citep{fernandez2023stablesignaturerootingwatermarks}}
        & \multicolumn{2}{c}{\textbf{VideoSeal}~\citep{fernandez2024videosealopenefficient}} \\
        \cmidrule(lr){2-3}
        \cmidrule(lr){4-5}
        \cmidrule(lr){6-7}
        \textbf{Dataset}
        & Mean & $>1.5\alpha$
        & Mean & $>1.5\alpha$
        & Mean & $>1.5\alpha$ \\
        \midrule

        \multicolumn{7}{@{}l}{\textbf{\boldmath $\alpha = 1\%$}} \\
        COCO (held out) & 1.01 & 0.2 & 1.00 & 0.0 & 1.00 & 0.0 \\
        SD~2.1 (generated) & 1.01 & 0.3 & 0.94 & 15.1 & 0.99 & 0.4 \\
        DTD & 1.10 & 5.2 & 1.23 & \textbf{26.4} & 1.00 & 0.6 \\
        Flowers102 & 0.91 & 10.9 & 0.92 & 18.6 & 1.00 & 0.1 \\
        Oxford-IIIT Pets & 0.97 & 0.6 & 0.92 & 13.4 & 1.01 & 0.3 \\
        WebSight & 1.31 & 30.4 & 0.98 & 16.4 & 1.00 & 3.2 \\
        Chest X-ray & 1.52 & \textbf{40.1} & 0.69 & 10.6 & 0.99 & \textbf{11.7} \\
        \midrule

        \multicolumn{7}{@{}l}{\textbf{\boldmath $\alpha = 5\%$}} \\
        COCO (held out) & 5.02 & 0.0 & 5.01 & 0.0 & 5.02 & 0.0 \\
        SD~2.1 (generated) & 5.04 & 0.0 & 4.87 & 14.0 & 4.97 & 0.0 \\
        DTD & 5.22 & 0.0 & 5.61 & \textbf{19.4} & 5.01 & 0.0 \\
        Flowers102 & 4.76 & 6.9 & 4.81 & 19.0 & 5.01 & 0.0 \\
        Oxford-IIIT Pets & 4.86 & 0.0 & 4.81 & 12.4 & 5.01 & 0.0 \\
        WebSight & 5.45 & 20.1 & 4.97 & 18.6 & 5.01 & 0.5 \\
        Chest X-ray & 6.33 & \textbf{31.3} & 4.28 & 16.8 & 4.99 & \textbf{6.6} \\
        \bottomrule
    \end{tabular*}
\end{table*}

\begin{figure}[t]
    \centering
    \includegraphics[width=\linewidth]{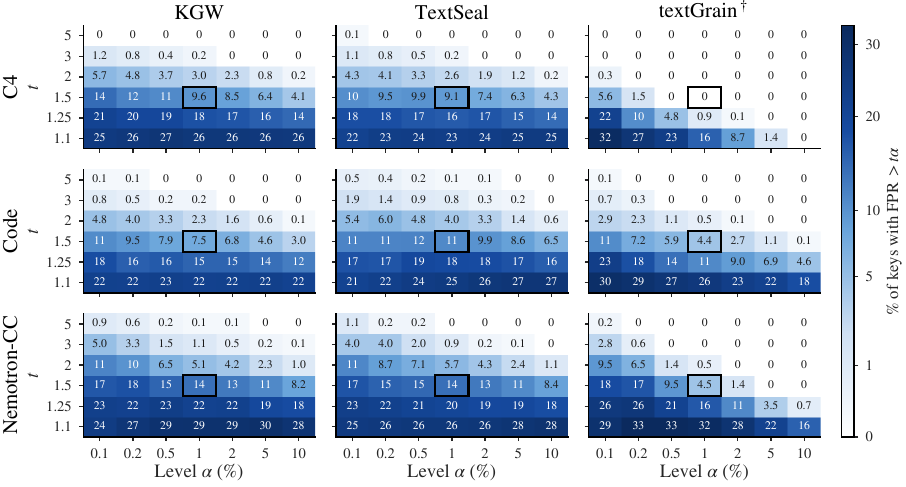}
    \caption{\textbf{Text watermarks have bad keys at every level.} Each cell is the percentage of $1{,}000$ keys whose FPR exceeds $t\alpha$ at the theoretical cutoff. Horizontal axis: level $\alpha$. Vertical axis: threshold $t$. The outlined cell, $\alpha = 1\%$ and $t = 1.5$, is the setting used in the rest of the paper. $^\dagger$Our reimplementation.}
    \label{fig:alpha_grid_text}
\end{figure}

\begin{figure}[t]
    \centering
    \includegraphics[width=\linewidth]{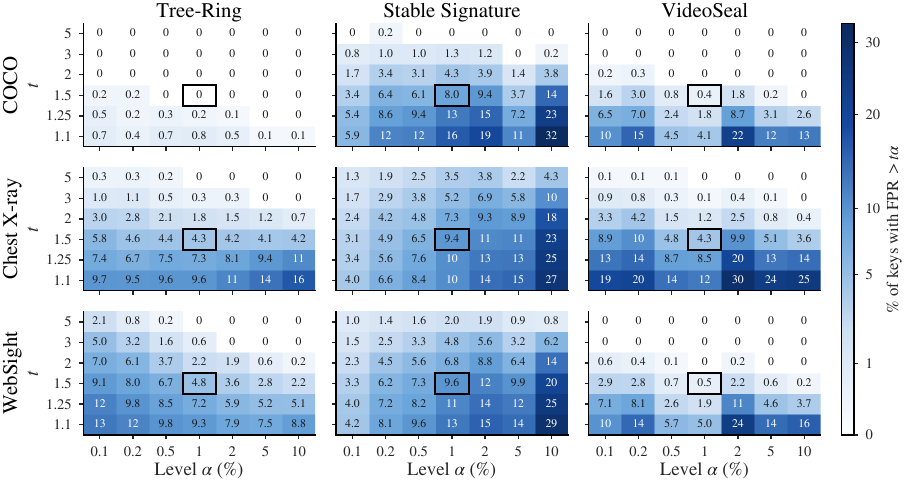}
    \caption{\textbf{Stable Signature has bad keys at every level, most at large levels.} As in \cref{fig:alpha_grid_text}, for the image watermarks on held-out COCO, chest X-rays, and WebSight.}
    \label{fig:alpha_grid_image}
\end{figure}

\end{document}